# High-fidelity Multiphysics Modelling for Rapid Predictions Using Physics-informed Parallel Neural Operator

Biao Yuan[a]  
cnby@leeds.ac.uk

He Wang[b]  
he_wang@ucl.ac.uk

Yanjie Song[a]  
cnys@leeds.ac.uk

Ana Heitor[a]  
A.Heitor@leeds.ac.uk

Xiaohui Chen[a*], Corresponding author  
X.Chen@leeds.ac.uk

a. Geomodelling and Artificial Intelligence Centre, School of Civil Engineering, University of Leeds, Leeds, LS2 9JT, UK

b. UCL Centre for Artificial Intelligence, Department of Computer Science, University College London, London, WC1E 6EA, UK

**Abstract**

Modelling complex multiphysics systems governed by nonlinear and strongly coupled partial differential equations (PDEs) is a cornerstone in computational science and engineering. However, it remains a formidable challenge for traditional numerical solvers due to high computational cost, making them impractical for large-scale applications. Neural operators' reliance on data-driven training limits their applicability in real-world scenarios, as data is often scarce or expensive to obtain. Here, we propose a novel paradigm, physics-informed parallel neural operator (PIPNO), a scalable and unsupervised learning framework that enables data-free PDE modelling by leveraging only governing physical laws. The parallel kernel integration design, incorporating ensemble learning, significantly enhances both compatibility and computational efficiency, enabling scalable operator learning for nonlinear and strongly coupled PDEs. PIPNO efficiently captures nonlinear operator mappings across diverse physics, including geotechnical engineering, material science, electromagnetism, quantum mechanics, and fluid dynamics. The proposed method achieves high-fidelity and rapid predictions, outperforming existing operator learning approaches in modelling nonlinear and strongly coupled multiphysics systems. Therefore, PIPNO offers a powerful alternative to conventional solvers, broadening the applicability of neural operators for multiphysics modelling while ensuring efficiency, robustness, and scalability.

## 1. Introduction

The modelling of various physical and complex phenomena is required in many scientific and engineering fields. Partial differential equations (PDEs) are arguably the most widely used mathematical tools for such purposes in several disciplines, including finance, wave dynamics, heat transfer, solid mechanics, fluid mechanics, quantum mechanics and electromagnetism(1-4), and their applications in geological and geotechnical engineering, biomedical engineering, chemical engineering, environmental engineering and materials engineering(4, 5). Therefore, it is essential to solve PDEs rapidly and accurately to gain insight into the dynamics of the underlying physical systems to solve natural phenomena and engineering problems. Nonetheless, analytical solutions to PDEs are seldom available, especially for complex physical systems. Therefore, various numerical methods have been developed to compute approximate solutions of PDEs. These include the finite volume method (FVM), finite difference method (FDM), boundary element method (BEM), spectral methods, and the finite element method (FEM), amongst others(6-11). While traditional methods have been proven effective in solving PDEs, they still face limitations in terms of computational dimensions and complexity, as well as challenges in terms of computational convergence and inverse problem-solving. This often results in high computational costs(6-10, 12). In recent years, the emergence of artificial intelligence (AI) has brought about a new paradigm for solving PDEs, which has already attracted considerable interest from researchers in mathematics and computational physics(13, 14). Despite the effectiveness of purely data-driven neural networks in approximating PDE solutions, their results remain susceptible to significant errors. Moreover, there is no guarantee that the learning process will satisfy the underlying physical laws. To address these shortcomings, physics-informed neural networks (PINNs) are being used to solve complex problems that can be discretised with PDEs(15-17). PINNs effectively combine the strengths of deep learning with physical principles. Indeed, past studies have proven that PINNs are effective in solving various complex problems across multiple fields, particularly those involving strong nonlinearities, multiphase flows, turbulence, and surge waves(15). Given neural networks are universal function approximators, PINNs can serve as an alternative model for modelling nonlinear systems, especially in forward inference and inverse problems. These inverse problems include tasks such as parameter identification, flow field reconstruction, and physical laws discovery, where PINNs often outperform traditional methods(18). Furthermore, PINNs are well-suited for solving high-dimensional problems, which is mainly due to the incorporation of Monte Carlo methods in the construction of loss functions(19). From a mathematical standpoint, PINNs can be classified into two formats: the strong form(17) and the energy form(20, 21). A variety of neural network architectures are extensively utilized for solving PDEs, for instance Fully Connected Neural Networks (FC-NNs)(17, 18), Recurrent Neural Networks (RNNs)(22-24), Convolutional Neural Networks (CNNs)(25-27), Graph Neural Networks (GNNs)(28, 29), Generative Adversarial Networks (GANs)(30), Extreme Learning Machine (ELM)(31, 32), and Kolmogorov-Arnold Networks (KANs)(33-35). The primary challenge facing neural network-based solvers is achieving high performance in solving diverse and complex PDE problems. To tackle this, in past studies a range of enhancement techniques have been proposed, for instance adaptive training weights, adaptive activation functions, adaptive sampling methods, domain decomposition, transfer learning, and loss function optimization(15, 36). These approaches have significantly improved the capabilities of PINNs in scientific computing.

Despite the availability of powerful solvers for PDEs, conventional numerical techniques and PINNs still face challenges related to convergence and accuracy(37). In addition, these methods are limited to solving one type of problem at a time, which means that recomputation is necessary whenever input variables change. This highlights the ongoing need for more generalized and robust models. PINNs specifically learn the mapping between two finite-dimensional vector spaces, and this can be mathematically extended to the mapping between two infinite-dimensional function spaces. As a result, neural operators(38) have been proposed, which are purely data-driven originally. The most basic examples of these neural operators include DeepONet(39-42), Fourier Neural Operator (FNO)(43, 44), Graphical Neural Operator (GNO), Convolutional Neural Operator (CNO)(45, 46), Wavelet Neural Operator (WNO)(47, 48), and Laplace Neural Operator (LNO)(49), et.al. Among them, DeepONet has gained significant attention due to its excellent approximation performance and robustness(50). FNO introduced by Li et al.(43), uses Fourier filtering to enhance the model's training speed and generalization while reducing the risk of overfitting. Moreover, FNO can learn from low-resolution data and infer high-resolution physical fields. The same authors of FNO also applied GNO to learn the relationships among field variables, revealing that the performance of the FNO surpasses that of the GNO(51). However, the effectiveness of the FNO is limited by complex boundary conditions. The fundamental operational principle of neural operators is to utilise neural networks for learning function mappings or operators between inputs and outputs(52). Kernel integration algorithms constructed using the neural operator layer can approximate arbitrary continuous linear or nonlinear operators. The efficiently trained framework can learn solutions to PDEs for different input scenarios, which is not currently possible with traditional data-driven and physics-informed neural networks. To illustrate, the input functions may be the initial conditions, boundary conditions, parameter conditions or forcing source terms of the PDEs, while the output is the objective function field of interest, and vice versa. Moreover, neural operators frequently approximate continuous functions with a restricted number of discrete inputs, underscoring the significance of discrete invariance and universal approximation theorem for them, which have been demonstrated to exist(39). Although neural operators outperform traditional grid-based methods for high-dimensional problems, they typically require high-quality training data and are computationally demanding(53). To effectively reduce data dependency, physics-informed deep operator networks (PI-DeepONet) have been introduced as innovative solutions that enhance the performance and reliability of modelling by incorporating essential physical constraints(54-57). This breakthrough minimises reliance on ground truth data, yet it still faces challenges in achieving significant performance improvements(50, 58). The evolution of this approach leads to the development of the physics-informed neural operator (PINO), which refines the operator learning process(53, 59). The PINO begins by training with a large dataset and subsequently fine-tunes the predicted solutions using physical equations applied to a test dataset, resulting in high accuracy. The neural operator method stands out for its remarkable ability to learn solutions to a wide range of parametric PDEs. In this context, a single operator can efficiently map the input function space to the corresponding solution space. Its successful application across diverse fields—such as multi-scale modelling, multi-physics simulations, climate modelling, and inverse design—demonstrates its potential as a robust modelling tool for complex physical systems(60). Despite these advancements, current operator learning studies still rely on significant amounts of data. Even those grounded in physics depend on traditional PDE solvers for their training processes. When removing real data and retaining only physical laws, they fail in certain problems, suggesting that the original neural operators remain data-driven. Therefore, reducing this dependency is crucial to improving generalisation accuracy and robustness, as well as preventing failures in accurately capturing authentic physical systems. Furthermore, the need for large training datasets poses a formidable barrier, especially when data generation is costly or inaccessible, thereby limiting the potential for innovation and development in this area.

This study proposes a novel approach for multiphysics modelling, namely physics-informed parallel neural operator (PIPNO), that can be used in different scenarios where data is difficult to obtain. The main findings and contributions of this study are summarised below. (1) PIPNO incorporates ensemble learning into the design of the parallel kernel integration algorithm within function spaces, improving both compatibility and computational efficiency for multiphysics modelling with nonlinear and strongly coupled PDEs. (2) The approach is scalable and based on unsupervised learning, whereby the governing physics itself can be used as the source of knowledge, rather than relying on ground truth data. Thus, PIPNO can make reasonable and accurate predictions for the solution functions of PDEs without the need for any assumptions or prior knowledge. (3) PIPNO efficiently achieves high-fidelity and rapid predictions for complex multiphysics systems. (4) Its performance in terms of generalisation accuracy and robustness is demonstrably superior to that of existing state-of-the-art research. (5) The efficacy of the proposed method has been demonstrated in this study through six illustrative examples, which are commonly employed to simulate geotechnical settlement problems, material phase separation problems, electromagnetic field modelling, quantum mechanics, fluid dynamics and water flow engineering. In quantitative terms, PIPNO is approximately 100 times faster than conventional PDE solvers and PINNs in terms of inference speed. With powerful learning capabilities, the proposed method has the potential to accelerate scientific exploration in computational mechanics, computational physics, computational engineering, and other fields, and to advance the creation of corresponding large-scale AI models.

The paper is structured into several sections. Section 2 outlines the problem of solving PDEs in multiphysics systems using neural operators. The framework, principles, and operator learning process of the proposed method are introduced in Section 3. Then, Section 4 presents the numerical experiments conducted to evaluate the performance of the proposed models against the baseline methods. Finally, the discussion and conclusion are summarised in Section 5.

## 2. Problem statement

Partial Differential Equations (PDEs) are a class of equations that involve partial derivatives of unknown functions(2). They are widely used to describe various phenomena in nature and engineering, including heat transfer, fluid flow, wave dynamics, elastodynamics, and electromagnetic fields, among other multidisciplinary issues. Specifically, parametric PDEs depend not only on the standard time and space variables but also on additional parameters, such as physical properties, geometry, or initial conditions. These parameters can influence the coefficients, initial conditions, boundary conditions, or source terms of the equation. In this study, the general form of PDEs that depicts the multiphysics system is considered as follows:

$$\begin{cases} \mathcal{N}_1(t, x_1, x_2, \dots, x_n, u, \frac{\partial u}{\partial t}, \frac{\partial u}{\partial x_1}, \frac{\partial u}{\partial x_2}, \dots, \frac{\partial^2 u}{\partial x_1^2}, \dots, \mu) = 0, x \in \Omega, t \in [0, T], \mu \in \mathcal{P} \\ \mathcal{N}_2 = 0; \dots; \mathcal{N}_i = 0 \end{cases} \quad (1)$$

$$u(0, x_1, x_2, \dots, x_n) = u_0(x_1, x_2, \dots, x_n), x \in \Omega, t = 0 \quad (2)$$

and

$$\mathcal{B}\left[u, \frac{\partial u}{\partial x_1}, \dots\right] = u_b(t, x_1, x_2, \dots, x_n), x \in \partial\Omega, t \in [0, T] \quad (3)$$

Where $x = (x_1, x_2, \dots, x_n)$ is the independent variable in the spatial domain $x \in \Omega$ and $\boldsymbol{u} = [u_1(t, x_1, x_2, \dots, x_n), u_2(t, x_1, x_2, \dots, x_n), \dots]$ is the unknown functions in the PDEs that need to be solved in the spatial domain $x \in \Omega$ and the temporal domain $t \in [0, T]$; $\frac{\partial u}{\partial t}$ is the time derivative of $u$ with respect to $t$; $\frac{\partial^n u}{\partial x_1^n}$ is the spatial differential operator with the order of $n$ acting on $u$ with respect to $x$. $\mu$ represents the parameter vector and $\mathcal{P}$ is the parameter space, e.g., material properties, geometry, etc. $\mathcal{N}$ is the partial differential operator, which represents the function mapping of the whole PDEs system and is also the research object in this paper. In particular, if a physical system is described by multiple physical laws or coupling relationships expressed through several PDEs, then this set of equations includes multiple $\mathcal{N}$. The initial condition of the PDEs system is set to $u_0(x_1, x_2, \dots, x_n)$ and the boundary condition is considered as $\mathcal{B}\left[u, \frac{\partial u}{\partial x_1}, \dots\right] = u_b(t, x_1, x_2, \dots, x_n)$, where $u_b$ is a specifically defined boundary function and $x \in \partial\Omega$ indicates the boundary of the spatial domain.

The primary objective of this paper is to design an innovative type of physics-informed neural operator that learns the relationship between initial functions and solution functions of PDEs. This is achieved primarily by analysing their physical dynamics as defined by the governing equations. The PDEs that control the physical system, the initial conditions as well as the boundary conditions are the only known and exact preconditions, and there is no request for any labelled data or data-driven learning. The function mapping $G$ can be expressed as:

$$G: a(t, x_1, x_2, \ldots, x_n) \xrightarrow{Mapping} [u_1(t, x_1, x_2, \ldots, x_n), u_2(t, x_1, x_2, \ldots, x_n), \ldots] = \boldsymbol{u} \qquad (4)$$

Where $a(t, x_1, x_2, \ldots, x_n)$ is the initial function input into the physics-informed neural operators as a training variable used to learn the mapping to the solution functions for the PDEs system. Previous studies have reported that neural networks can approximate any continuous operator(39). It is important to note that physics-informed neural operators differ from the originally proposed PINNs. While PINNs can only solve specific problems that means any change on initial conditions, boundary conditions, geometry, or materials requires retraining, physics-informed neural operators learn a series of mappings to the family of PDEs. As a result, they can quickly obtain the target solution even when these conditions change, making it one of the most promising developments in the field of computational mechanics. The original concept of operator learning was primarily data-driven, making it highly suitable for problems involving large volumes of data. However, vanilla operator learning can face challenges due to ambiguity in physics and may struggle with insufficient data volumes, which can be either costly or unavailable. At present, most of the research on operator learning still relies on substantial amounts of data, even on physics-informed operator learning, significantly limiting its performance and development.

Therefore, another goal of this research is to develop neural operators that are entirely driven by physical principles, specifically for use in multiphysics systems. This approach enables unsupervised learning without the need for data and enhances accuracy and generalisation robustness. To achieve this, it is necessary to create frameworks that possess advanced physical learning capabilities. The entire neural operator training process is unsupervised and does not require any labelled data, only the initial conditions and boundary conditions, as well as the exact PDEs system are given. By deciphering and integrating the physical information of a system of PDEs, the solution of the PDEs system in the entire domain is sequentially mapped among the whole domain from the given initial condition as the exact physical state at the first time point. The neural operator training process's loss function is constrained by physical laws. The primary goal is to approximate the solution of the PDEs system by training neural operators to optimise a loss function composed of PDEs residuals. These methods can act as numerical solvers and surrogate models for parametric PDEs with only exact physics information given and have the potential to address the corresponding inverse problems and as cornerstones for the next generation of large-scale AI models for computational science.

### 3. Methodology

Physics-informed parallel neural operator (PIPNO) aims to learn the mappings between infinite-dimensional input and output function spaces. This study focuses on learning solutions to PDEs based on specific input conditions. The objective is to design a physics-informed operator learning framework that outperforms existing state-of-the-art methods while being independent of data. The following sections will provide a detailed description of the framework's structure, along with a high-level overview of the kernel integration algorithm that supports this approach and the mathematical principles of the universal approximation theorem. Moreover, since the physics-informed parallel operator learning directly learns the output fields from exact physical information, it is essential to incorporate residuals of the governing equations into the loss function. This often requires computing derivatives, constructing the loss function, and fine designing the optimisation algorithm. In this section, the methodology of this study will be discussed in detail.

### 3.1 Physics-informed parallel neural operator

The fundamental structure and framework of the proposed method and its core working principles is outlined in the schematic diagram which is shown in Figure 1. It can be observed that the PIPNO framework receives initial function as input and produces the output state for the physical systems. Based on the previous problem statement, the objectives to be achieved by the process are outlined as follows:

$$\hat{u} = G_{\theta_{PNO}^*}(u)(a, x) \tag{5}$$

Where $\hat{u} = [\hat{u}_1, \hat{u}_2, \ldots, \hat{u}_j]$ represents the approximate solutions of PDEs in a complex physical systems surrogated by PIPNO, and $\hat{u}_j$ denotes different physical variables. $a = a_0(t, x_1, x_2, \ldots, x_k)$ is the initial function for function to function operator mapping and $x = (t, x_1, x_2, \ldots, x_k)$ is the spatiotemporal coordinates. $G_{\theta_{PNO}^*}(u)$ indicates the operator acting on the physical variable $u$, while $\theta_{PNO}^*$ is the network parameter to be optimised in the proposed framework.

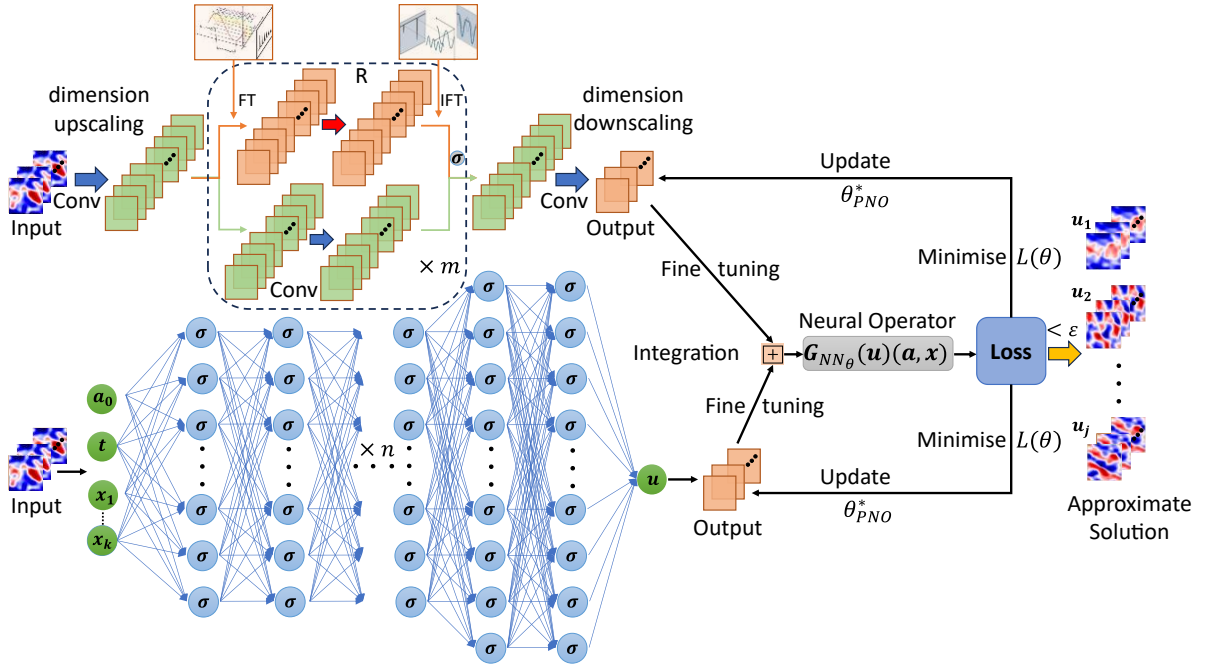

**Figure 1** The functional architecture of the PIPNO with inputs, parallel operators, physics-loss optimisation and outputs for PDEs in multiphysics systems with spatial and temporal coordinates $(t, x_1, x_2, \ldots, x_k)$. The initial function for physics-informed operator learning is $a_0$ in the inputs. The architecture is designed to learn the mapping from the input function to the solution of the PDEs. The inputs are processed by independent parallel neural operators, which transform them into higher-dimensional spaces. In Banach space, the hidden operators

in the PDEs are learned using multilayer perceptrons. Meanwhile, in Hilbert space, the inputs undergo both Fourier and physical convolutions; Here, FT refers to the Fourier Transform and IFT refers to the Inverse Fourier Transform, R is the linear transform or convolution for frequency filtering in the Fourier space. Ultimately, the dimensions are reduced back to the physical space, and the outputs of the parallel neural operators are weighted and integrated to derive solutions for the target PDEs. These solutions must satisfy the specified PDE, boundary conditions (BC), and initial condition (IC) loss constraints. The nonlinear activation function $\sigma$ is the Gaussian Error Linear Unit (GeLU). Additionally, numerical differentiation is employed to compute the partial derivatives in the governing equations. $\theta_{PNO}^*$ represents the parameters of the framework that need to be optimised.

The whole architecture is designed to learn mappings from arbitrary functions to solutions of PDEs. Incorporating the advantages of integrated learning is an important feature of the proposed framework, and the complementary nature of parallel operator learning improves the generalisation accuracy and robustness. This is inspired by the operator theory in classical generalised functional analysis, where the essence of neural operator approximation is the design of the kernel integration algorithm. It can be learnt from Figure 1 that in the upper parallel neural operator, the inputs are convolved to the high-dimensional space, and the intermediate result is transformed to the Fourier space for linear transformation. Therefore, the mapping relationship between functions can be learnt in the Fourier space, and the result is converted back to the physical space by inverse Fourier transform. Here, the kernel integration algorithm based on the Hilbert space is utilized to approximate the operator, and the idea of the residual network (ResNet) is employed so that the convolution is also performed directly on the intermediate result after dimension upscaling. Finally, the results of these two are added together. At this point, this parallel operator path acts as a specific class of Fourier neural operator (FNO), and there are $m$ Fourier blocks in total. In addition, the inputs are simultaneously fed to a serial of multilayer perceptrons in the lower parallel neural operator, where the number of layers is $n$ in total. The functional mapping relationship between the inputs and outputs is learnt based on the universal approximation theorem. Eventually, the outputs of two parallel neural operators are weighted and integrated to obtain the approximate solution to the target PDEs. In this context, the inner product integration similar to that used in DeepONet has been considered. However, several numerical tests have indicated that linear weighting integration is the most effective approach. This can also be explained through mathematical principles: the inner product signifies that the two networks are interconnected or coupled, while linear weighting illustrates the independence between the neural operators. Furthermore, the approximate solutions must adhere to the relevant physical laws, which are surrogated by the constructed loss function that incorporates constraints for initial conditions, boundary conditions, and PDEs. In the PDE loss of the governing equations, the partial derivatives are computed using numerical differentiation methods such as Fourier difference or finite difference. As previously mentioned, the key aspect of physics-informed parallel operator learning is the design of suitable and efficient kernel integration algorithms, as well as the approximation of these kernels and integrals using the computational principles of neural networks. Specifically, in the lower parallel neural operator, the kernel integration algorithm closely resembles the Taylor series expansion based on Banach spaces. Meanwhile, the kernel integration algorithm in the upper parallel neural operator is derived from the Fourier transform within a more specific Hilbert space.

### 3.2 Function spaces and parallel operator learning

This study achieves mapping between infinite dimensional input and output function spaces by designing specific neural operator paradigms dominated by the laws of physics. Specifically, for the input function $a(x,t) \in \mathcal{A}$ and the output function $u(x,t) \in \mathcal{U}$, the objective is to study the solution mapping operator of PDEs, $\mathcal{G}: \mathcal{A} \to \mathcal{U}$, where $x \in \Omega$ and $t \in \mathbb{R}^+$. $\Omega \subset \mathbb{R}^d$ is the $n-$dimensional domain with boundary $\partial\Omega$. In addition, $\mathcal{A}$ and $\mathcal{U}$ are spaces of functions defining two complete normed linear spaces with specific partial derivatives on the domain $\mathbb{R}^{d_a}$ and the domain $\mathbb{R}^{d_u}$ respectively, these function spaces are usually called Banach spaces $\mathcal{B}$. In mathematical physics, the Green's function is a powerful tool for solving linear PDEs. Its general form is as follows:

$$u(x) = \int_\Omega G(x,x')f(x')dx' \tag{6}$$

Where $u(x) \in \mathcal{U}$ is the solution and $f(x') \in \mathcal{A}$ is the source term of the PDE, $G(x,x') \in \mathbb{C}$ is the Green's function to describe the role of partial differential operators, and $x$ and $x'$ are the variables that indicate two locations or points in the spatiotemporal domain in the problem. Inspired by Green's function, it is possible to generalise the source term $f(x') \in \mathcal{A}$ to the case of arbitrary input function $a(x') \in \mathcal{A}$ and to generalise linear PDEs to nonlinear PDEs. This allows the design of kernel integration algorithms of neural operators for solving complex PDEs. Therefore, for any nonlinear PDEs, the above integral equation can be generalised to an Urysohn-type integral equation:

$$u(x) = \int_\Omega \mathscr{k}(x,\delta)f(\delta,a(\delta))d\delta + g(x) \tag{7}$$

In this context, $\mathscr{k}(x,\delta) \in \mathbb{C}$ refers to the kernel of the nonlinear integral equation that represents the nonlinear counterpart of the linear component $G(x,x')$ of the Green's function. The symbol $g(x)$ denotes an exact kind of linear transformation. This nonlinear integral equation serves as a generalization of the Green's function and is applicable for any arbitrary input function within the neural operator framework. And the input space is lifted to a higher dimensional space in the neural operator before executing the above kernel integration algorithm. Firstly, the lower parallel neural operator mentioned above, called $Pnet_1$. It has the similar universal approximation theorem as PINNs and DeepONet, making its mathematical structure similar to the Taylor series:

$$u(x) = \int_\Omega \mathscr{k}_t(x,\delta)f(\delta,a(\delta))d\delta \approx \sum_{i=1}^n \varphi_i(x)\alpha_i(u) \tag{8}$$

Where $\varphi_i(x)$ serves as a basis function in the Taylor series and is referred to as the Taylor kernel $\mathscr{k}_t(x,\delta)$ in kernel integration algorithms for the lower parallel path, and used to approximate the mapping between input and output function spaces in Banach spaces. This method utilises the neural network to fit the objective function derived from the coordinate and input spaces. Instead of directly fitting the objective function, it fits the basis or kernel function, so the process is akin to function approximation in numerical analysis. Unlike traditional algorithms that necessitate pre-selection of basis functions, this approach allows the neural network to adaptively select suitable basis functions and weights based on the available data. Similarly, $Pnet_2$ is the upper parallel neural operator involving Fourier kernels $\Bbbk_f(x,\delta)$.

$$u(x) = \int_\Omega \Bbbk_f(x,\delta)\, f(\delta, a(\delta))d\delta + g(x) \approx \sum_{i=1}^{m} (\Bbbk_i(x)v_i(x) + \mathcal{T}v_i(x)) \tag{9}$$

There are $m$ layers of neural networks in the upper parallel path, and the very first input function is $a(x)$. $\Bbbk_i(x)$ and $v_i(x)$ represent the kernel or basis function and provisional output of the $i-$th layer of the network, respectively. $\mathcal{T}$ indicates the linear transformation in the Urysohn-type integral equation, which is expressed by a linear convolution in the neural operator. Therefore, the output function $v(x)$ can be evaluated at any point, while the input function $a(x)$ can be mapped at any resolution.

$$v(x) = \sigma\big(\mathcal{K}(v)(x)\big) = \sigma\left(\int \Bbbk(x,\delta)\, v(x)dx + \mathcal{T}v(x)\right) \approx \sum_{i=0}^{m} \sigma(\sum_{j=1}^{k} (\Bbbk_i(x_j)v_i(x_j)\Delta x_j + \mathcal{T}_j v_i(x_j)) \tag{10}$$

When $i = 0$, $v(x)$ is the initial space $a(x)$. $\sigma$ denotes the nonlinear activation function that brings nonlinearity to the feature mapping in neural operators. Moreover, the above process can be summarised using the kernel integral operator in the neural operator:

$$\mathcal{G}(v_i)(x) \coloneqq \sigma\big((\Bbbk(a,\theta) \cdot v_i)(x) + \mathcal{T}v_i(x)\big); x \in \Omega, i \in [0,m] \tag{11}$$

Where $\Bbbk(a,\theta)$ is the kernel function learned by the neural operator, $a$ represents the input function space, and $\theta$ indicates the parameter of the neural operator. As mentioned before, the upper parallel path contains a class of kernels with time-frequency transformations. This involves a more specific function space known as Hilbert space $\mathcal{H}$. Firstly, $v(x)$ can be transformed to $\hat{v}(k)$ in the frequency domain by $\hat{v}(k) = \int_\mathcal{H} v(x)e^{-i2\pi kx}dx$. In the Fourier space, feature learning can be achieved by using a low-pass frequency filter $\mathcal{R}_\theta$ to achieve convolution in frequency-domain space. Then, a new $v(x)$ is obtained by $v(x) = \int_\mathcal{H} \hat{v}(k)e^{i2\pi kx}dk$. Finally, the linear transformation $\mathcal{T}$ can be substituted by convolution $(t * v)(x) = \int_\mathcal{B} t(\tau)v(x-\tau)d\tau$. In this context, the kernel in Equation (11) can be expressed as $(\Bbbk(a,\theta) \cdot v_i)(x) = \mathcal{F}^{-1}\big(\mathcal{R}_\theta * \mathcal{F}(v_i)\big)(x)$. With the above definition of parallel neural operators, the different neural operators $Pnet_1$ and $Pnet_2$ are made parallel using linear weighting integration:

$$\xi_1 Pnet_1 + \xi_2 Pnet_2 = \xi_1 \int_\Omega \Bbbk_t(x,\delta)\, f(\delta, a(\delta))d\delta + \xi_2 \int_\Omega \Bbbk_f(x,\delta)f(\delta, a(\delta))d\delta + g(x) \tag{12}$$

Therefore, the parallel mapping operator implemented by the proposed framework in this study can be expressed as follows:

$$\hat{u}(x) = \mathcal{G}_\theta(u)(a,x) = \int (\xi_1 \mathcal{k}_t(x,\delta) + \xi_2 \mathcal{k}_f(x,\delta))f(\delta, a(\delta))d\delta + g(x) \tag{13}$$

$\hat{u}(x)$ is the approximate solution by the framework. $\xi_1$ and $\xi_2$ are the coordination weight factors of $Pnet_1$ and $Pnet_2$, respectively. The parallel kernel is ultimately shown below.

$$\mathcal{K}(a)(x) = \xi_1 \mathcal{k}_t(x,\delta) + \xi_2 \mathcal{k}_f(x,\delta) = \xi_1 \varphi_i(x) + \xi_2 e^{e^{-i2\pi kx}} \tag{14}$$

$\mathcal{K}(a)(x)$ is the final basis or kernel function for physics-informed parallel neural operator. $\xi_1 : \xi_2$ can be recommended as 2: 1. Since the infinite dimensional space of the integral in Equation (13) cannot be defined in the neural operator, it is approximated by a finite dimensional parameterised space obtained by discretising the solution domain. Meanwhile, The output of the framework must meet the minimum error limit requirements, and the discretisation in $Pnet_1$ is described below.

$$\left\| \mathcal{G}_\theta(u)(a,x) - \sum_{i=1}^{n} \sigma(\varphi_i(x)\alpha_i(u)) \right\|_2 < \varepsilon_1 \tag{15}$$

In which $\mathcal{G}_\theta(u)(a,x)$ denotes the true operator need to be studied by the framework. The operator approximation is surrogated by $Pnet_1$ and the process can be expressed by $\sum_{i=1}^{n}\sigma(\varphi_i(x)\alpha_i(u)) = (W_n \cdot \sigma(\cdots \sigma(W_2 \cdot \sigma(W_1 \cdot [t,x,a]^T + b_1) + b_2)\cdots) + b_n)$, $W$ and $b$ are the weights and biases of neural networks, and $\varepsilon$ is the given error limit. In other words, the equation above represents the error formula that the neural operator must satisfy. In $Pnet_1$, it is assumed that $\mathcal{B}$ is a Banach space, while $\mathcal{M}_1 \subset \mathcal{B}$ and $\Omega \subset \mathbb{R}^d$ are two compact sets, respectively. $\mathcal{A}$ is another compact set in $\mathcal{C}(\mathcal{M}_1)$. $\mathcal{G}_\theta(u)(a,x): \mathcal{A} \to \mathcal{C}(\mathcal{M}_1)$ is the nonlinear continuous operator that the neural operator will approximate. Therefore, there will exist an $Pnet_1: \mathcal{A} \to \mathbb{R}$ with suitable network parameters $W$ and $b$, positive integer $n$, $t \in \Omega, x \in \mathcal{M}_1$ that satisfies the above error formula for any $\varepsilon_1 > 0$. Similarly, $Pnet_2$ needs to satisfy the discretised relationship described in the following equation:

$$\left\| \mathcal{G}_\theta(u)(a,x) - (\sum_{q=1}^{m} \sigma((\mathcal{K}^q([t,x,a]^T; \varphi)v^q)(x) + \mathcal{T}^q v^q(x))) \right\|_2 < \varepsilon_2 \tag{16}$$

In this parallel path, $m$ represents the total number of layers, while $q$ is the provisional output of a certain hidden layer. $\varphi$ represents the framework parameters and $\varepsilon_2$ is the tolerance error. This parallel neural operator approximates the mapping of operators between different function spaces in the Hilbert space $\mathcal{H}$ with smooth indices. If $\mathbb{L}^d$ is a domain with Lipschitz boundary, then the mapping $\mathcal{G}_\theta(u)(a,x)$ is between $\mathcal{H}^\mathcal{A}(\mathbb{L}^d, \mathbb{R}^d)$ and $\mathcal{H}^\mathcal{U}(\mathbb{L}^d, \mathbb{R}^d)$, and there will exist $Pnet_2: \mathcal{A} \to \mathbb{R}$ with appropriate learned kernel $\mathcal{K}^q$, parameter $\varphi$ and transform $\mathcal{T}^q$, positive integer $m, t \in \mathbb{R}^+, x \in \Omega$ that satisfies the above error limit for any $\varepsilon_2 > 0$. Overall, the framework must satisfy specific qualification to achieve the stated objective, as shown below:

$$\sup_{\mathcal{A}} \|\hat{\mathcal{G}}_\theta(u)(a,x) - \mathcal{G}^+(u)(a,x)\|_{\mathcal{U}} \leq \epsilon \tag{17}$$

Where $\hat{\mathcal{G}}_\theta(u)(a,x)$ denotes the operator of the parallel framework approximation, while $\mathcal{G}^+(u)(a,x)$ is the real operator. Meanwhile, the established tolerance error for the parallel neural operator framework is denoted as $\epsilon$. Thus, the framework aims to find the optimal path for approximating the operator mapping from input function space $\mathcal{A}$ to output function space $\mathcal{U}$ while satisfying the tolerance error.

### 3.3 Loss function and optimisation

In the proposed framework, the neural operator learns real physical scenarios quickly and accurately by relying solely on the constraints imposed by physical laws. This approach makes the entire framework unsupervised, eliminating the need for labelled data. Therefore, constructing an effective loss function is a crucial step in integrating the neural operator with physical information. Setting appropriate constraints for the proposed framework is essential to ensure the accurate approximation of PDEs. Accurate physical systems are typically represented by PDEs, which require complete preconditions, including initial conditions, boundary conditions, material parameters, external forcing source terms, and the PDEs that strictly describe the physical information. The core challenge lies in developing a loss function that captures the residuals of the PDE based on the control equations under specific conditions and parameters. Furthermore, this loss function can clearly and intuitively assess the model's performance. According to the control equations of the physical system, the PDE residual can be defined as follows:

$$\begin{cases} \mathcal{N}_1(t, x, \hat{u}, \theta) = 0 \\ \mathcal{N}_2(t, x, \hat{u}, \theta) = 0; \ldots; \mathcal{N}_i(t, x, \hat{u}, \theta) = 0 \end{cases} \tag{18}$$

Therefore, in a physical system described by a set of PDEs, denoted as $\mathcal{N}_1; \mathcal{N}_2; \ldots; \mathcal{N}_i$. There are $i$ PDEs in the physical systems in total, and each equation expresses fundamental laws of physics. The variables $(t, x)$ are the spatiotemporal coordinates of the domain in the problem statement, while $\hat{u}$ represents the approximate solutions to the PDEs generated by the neural operator. The symbol $\theta$ refers to the network parameters that help obtain these approximate solutions, which can be refined during the optimisation process of the loss function. It is essential to ensure that the approximate solutions provided by the neural operator are constrained by the original physical equations, thereby satisfying the equilibrium conditions inherent to those equations.

$$\theta = \arg\min_{\theta} Loss(\boldsymbol{G}_{\theta^*_{PNO}}(\boldsymbol{u})(\boldsymbol{a},\boldsymbol{x}), \boldsymbol{G}(\boldsymbol{u})(\boldsymbol{a},\boldsymbol{x})) \tag{19}$$

After obtaining the initial approximate output from the proposed framework for solving the PDEs, the results must be adjusted to meet a set of exact preconditions inherent to the physical system, such as the initial and boundary conditions, as indicated by the equation:

$$\mathcal{L}oss_c = \int_\Omega^{ic} \left|\widehat{U}_{ic}(x_k) - U_{ic}(x_k)\right|^2 dx + \int_{\partial\Omega}^{bc} \left|\widehat{U}_{bc}(x_k) - U_{bc}(x_k)\right|^2 dx \qquad (20)$$

The condition loss $\mathcal{L}oss_c$ includes the square integrations of initial condition and boundary condition constraints. Specifically, the initial condition loss depends on the input functions, such as initial functions at the beginning time, external forcing terms, coefficients, etc. In addition, The Mean Squared Error (MSE) discretisation is utilised instead of continuous integration in neural networks.

$$\mathcal{L}oss_c = MSE_c = \sum_\Omega \left\|\widehat{U}_{ic} - U_{ic}\right\|^2 + \sum_{\partial\Omega} \left\|\widehat{U}_{bc} - U_{bc}\right\|^2 \qquad (21)$$

Where $\widehat{U}_{ic}$ and $\widehat{U}_{bc}$ indicate the approximate values of ICs and BCs, while $U_{ic}$ and $U_{bc}$ are the true conditions, respectively. Therefore, the condition loss $\mathcal{L}oss_c$ can be computed as follows:

$$\mathcal{L}oss_c = \sum_{j=1}^{n_j} (\frac{1}{N_{ic}} \sum_{q=1}^{N_{ic}} \|\hat{u}_j - u_j\|^2) + \sum_{j=1}^{n_j} \sum_{k=1}^{n_k} (\frac{1}{N_{bc}} \sum_{q=1}^{N_{bc}} \left\|\mathcal{B}\left(\hat{u}_j, \frac{\partial \hat{u}_j}{\partial x_k}\right) - u_j^b\right\|^2) \qquad (22)$$

Similarly, the PDE loss $Loss_f$ summarises all PDE residual integrals over the domain of definition $\Omega$:

$$Loss_f = \int_\Omega^{pde} \left|\mathcal{N}_i(t, x, \widehat{U}(x_k), \theta)\right|^2 dx \qquad (23)$$

After MSE discretisation it can be expressed as:

$$\mathcal{L}oss_f = MSE_f = \sum_\Omega \left\|\mathcal{N}_i(t, x, \widehat{U}(x_k), \theta)\right\|^2 \qquad (24)$$

Ultimately, all PDE residuals in the domain can be computed as:

$$Loss_f = \sum_{i=1}^{n_i} \sum_{j=1}^{n_j} \sum_{k=1}^{n_k} (\frac{1}{N_f} \sum_{q=1}^{N_f} \left\|\mathcal{N}_i\left(t, x_k, \hat{u}_j, \frac{\partial \hat{u}_j}{\partial t}, \frac{\partial \hat{u}_j}{\partial x_k}, \ldots, \frac{\partial^n \hat{u}_j}{\partial x_k^n}, \ldots, \mu\right)\right\|^2) \qquad (25)$$

In physics-informed operator learning, the loss function serves as a crucial metric for evaluating the difference between predictions and the ground truth. The goal of constructing this loss function is to ensure that the MSE approaches zero by optimising the model's parameters through an appropriate optimisation algorithm. Finally, the total physical loss is defined as:

$$Loss = \alpha \cdot Loss_c + \beta \cdot Loss_f \qquad (26)$$

Where $\alpha = 5$ and $\beta = 2$ are the loss weights of each loss term. The choice of this set of weights is based on extensive experience with numerical experiments. It is important to note that the loss function requires computing derivatives on the outputs of the operator. Due to the intensive computation and inefficiency of automatic differentiation (AD), some alternatives can generally be used to compute differentials in neural operators. In this study, Fourier differentiation and finite differentiation are considered.

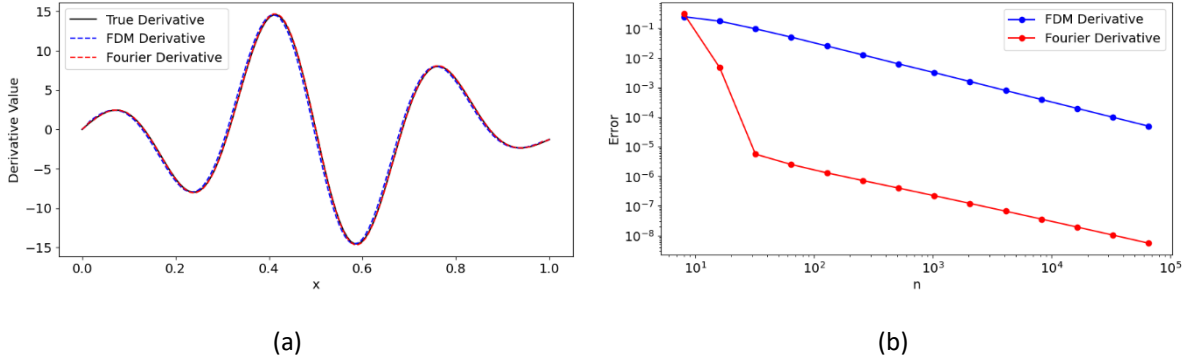

(a)                (b)

**Figure 2** The comparison of different numerical differentiations in the first-order derivative. (a) the comparison between true, forward Euler and Fourier derivatives when the number of intervals $n$=128; (b)the comparison of error convergence between forward Euler and Fourier derivatives when $n$ increases.

Moreover, Fourier differentiation is the priority, as it has been shown to have higher accuracy and faster error convergence than finite differentiation(11). The relevant results are also shown in Figure 2.

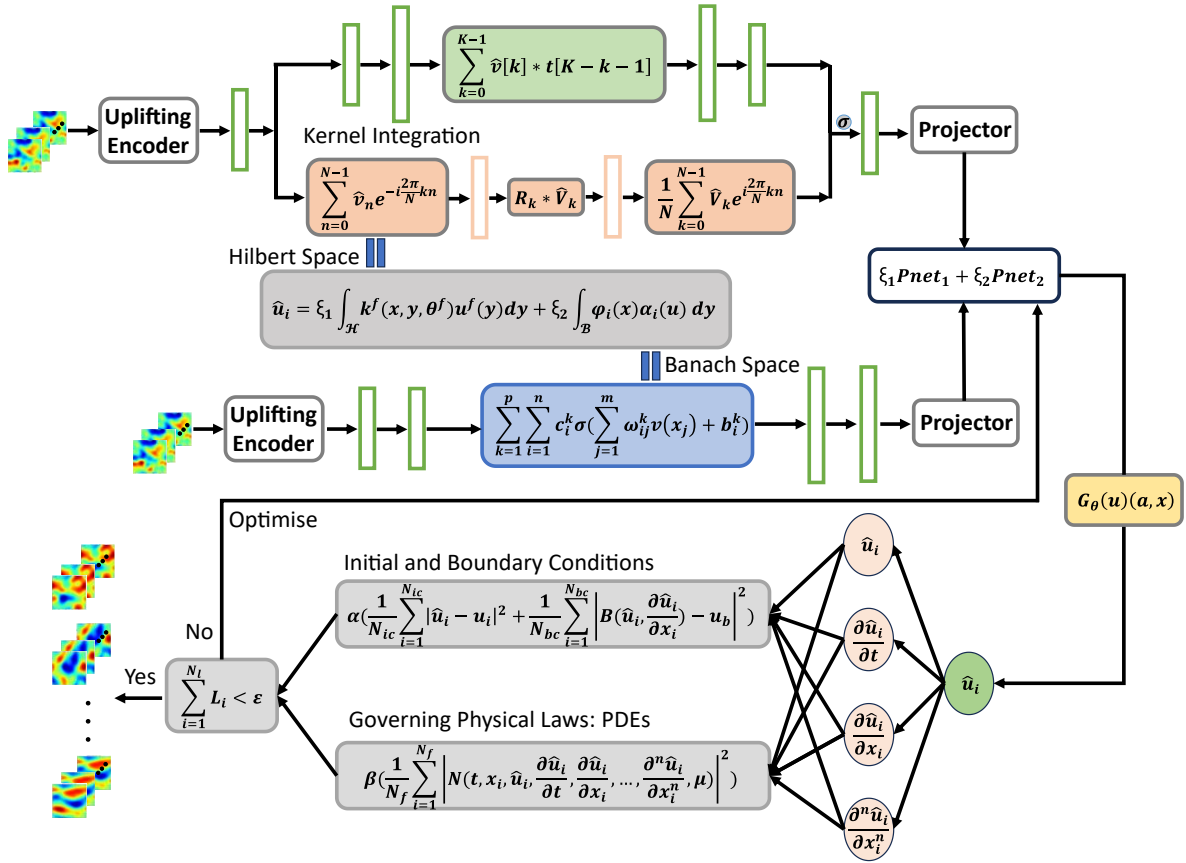

**Figure 3** The illustration of the learning process and optimisation principles for PIPNO.

With the previous theoretical foundations and methodology, the design of the kernel integration algorithm and the parallel operator learning theory relevant to this study are presented in full in Figure 3. Integrating the Taylor series with the Fourier kernel integration for ensemble learning is not unique to parallel neural operators. It is possible to couple different kernel integration operators or neural networks using the principle of complementarity of advantages. However, as mentioned earlier, the kernel-integral layer provides only frequency information, so this combination may fail to capture local features in the spatial domain. Nevertheless, the generalisation and robustness are significantly improved. On the other hand, ensemble operators pose a design challenge and require reasonable kernel integration algorithms to coordinate different operators. However, such methods are still potential alternatives for solving PDEs.

## 4. Numerical examples

To assesses the validity, accuracy, and robustness of the proposed framework, a series of benchmark PDEs are used as numerical examples related to multiphysics systems. The demonstration illustrates the advantages and potential of the proposed method for a diverse range of scientific and engineering applications. The numerical examples include several 1D and 2D linear and nonlinear PDEs in multiphysics systems: Consolidation equation; Allen-Cahn equation; Maxwell's equations; Schrödinger equation; Navier-Stokes equation; Shallow Water equations. Before the numerical experiments, various function samples with specific correlation and smoothness characteristics are generated for all numerical examples through the Matérn kernel of a Gaussian Random Field (GRF) as the input to the proposed framework. GRF is a random field employed to simulate spatially correlated structures, while the Matérn kernel is a flexible and widely utilized kernel function for describing the covariance structure of GRF. The specific parameter settings govern the fundamental attributes of the input function samples, such as the smoothness parameter $v$ dictating the differentiability, length scale $l$ controlling the range of correlations, and amplitude (variance) $\sigma$ of the generated field, which will be elaborated upon in the context of each distinct numerical case. Furthermore, numerical experiments reveal that initial conditions significantly influence accuracy and outcomes. Among these PDE physics systems, specific periodic boundary conditions are considered in all numerical scenarios. The first two PDEs have no coupling features and can be classified as single input-output mapping equations in PIPNO. Given that the Fourier transform is employed in the proposed framework, the periodic boundary conditions are inherently satisfied for the single physical variables, eliminating the necessity to incorporate supplementary boundary loss into the loss function. However, the last four PDEs are distinguished by explicit or implicit multi-physics field coupling, which has historically posed and is still a significant challenge for conventional numerical solutions, and the Fourier transform is only typically employed for a single input variable in the PIPNO, whereas the network outputs multiple variables. Thus, the boundary conditions cannot be constrained sufficiently, and additional and accurate boundary constraints need to be added to the loss function in this context. In this study, all the 1D numerical experiments are implemented in Pytorch and performed on the NVIDIA Tesla T4 GPU card, and 2D cases on the NVIDIA A100 GPU card.

The architectural configuration of the PIPNO framework is comprised of two distinct and independent parallel networks. In accordance with the illustrative examples, the structure of the first parallel network is set as follows: [$inputs, dimension\ upscaling, 64,64,64,64, dimension\ downscaling, outputs$], and the numbers of Fourier modes in PIPNO filters for 1D and 2D numerical examples are set as 36 and 12, respectively. While the second parallel network is structured as follows: [$inputs, 64,64,64,128,128, outputs$]. The ADAM algorithm with an initial learning rate of $1 \times 10^{-3}$ is used as the optimizer for PIPNO during the training process. A multi-step learning rate annealing algorithm is introduced during training to refine the convergence: the learning rate is reduced to a certain percentage of its original value when a predefined milestone is reached in a specified training period or epoch. This strategy allows the learning rate to be dynamically adjusted at critical stages of model training. For the one-dimensional PDE examples, the total number of epochs is 500, and the learning rate decays by 50% per 100 epochs, while for the two-dimensional PDE examples, the total number of epochs is 200, and the learning rate decays by 50% per 25 epochs. The neural operators are trained and optimised on the training datasets, and the prediction errors are evaluated on the testing datasets. The training and testing sets are completely different and have no intersection. $Ns$ and $Nt$ denote the number of training and testing samples, respectively, and the ratio is taken as 10:1. In addition, the numerical results indicate that more training samples lead to higher generalisation accuracy. However, due to the inherent limitations of the model, it will not increase indefinitely. Therefore, suitable and appropriate training and testing datasets are selected in this study. The batch sizes for the 1D training, 2D training and testing processes are 20, 10 and 1, respectively.

Furthermore, comparative studies of PIPNO under different physical systems are conducted and subjected to rigorous examination. The ground truth and the baseline algorithm are considered as references and a basis for comparison to evaluate the accuracy and effectiveness of the proposed method. In this study, the fourth-order Runge-Kutta method is employed to generate numerical solutions as references for all PDE numerical cases, except the Navier-Stokes equation, which is computed using the Fourier spectral method. PIFNO with a vanilla FNO structure as the backbone is used as the original baseline to facilitate comparisons with the proposed method. The reason for the choice is that it is the state-of-the-art method that has been widely demonstrated to be powerful and effective in solving PDEs at present. Moreover, elaborate error evaluation is required to quantify the difference between the predictions and references, the relative $L_2$ norm error (RL2E), mean absolute error (MAE), and root mean square error (RMSE) are introduced and described as follows:

$$\begin{cases} RL2E = \dfrac{\|\hat{u} - u\|_{L_2}}{\|u\|_{L_2}} \\ MAE = \dfrac{\sum|\hat{u} - u|}{N_f} \\ RMSE = \sqrt{\dfrac{\sum|\hat{u} - u|^2}{N_f}} \end{cases} \qquad (27)$$

Where $\hat{u}$ represents the framework predictions and $u$ is the ground truth. $\|\cdot\|_{L_2}$ denotes the computation of the $L_2$ norm, and $N_f$ is the number of points in the whole defined domain. Table 1 summarises the results of the numerical examples considered in this study, where the RL2E is taken as an error example. The data presented in the table illustrates the mean error on the testing dataset under different scenarios, along with its corresponding standard deviation. In order to facilitate the visualisation of the discrepancy in performance between the proposed method and the comparative baseline, the respective RL2E, MAE, and RMSE for different numerical examples are also presented in Figure 4.

**Table 1** Relative $L_2$ Errors of all the PDEs solution inference on testing datasets between different physics-informed neural operators and reference solution for all the numerical examples.

| PDEs | Spatiotemporal Resolution | Results | PIPNO | Vanilla PIFNO |
|---|---|---|---|---|
| Consolidation | $128 \times 100$ | RL2E | $2.82e-5 \pm 8.88e-7$ | $2.87e-4 \pm 5.01e-6$ |
| | | $Ns:Nt$ | $1000:100$ | $1000:100$ |
| | $64^2 \times 100$ | RL2E | $1.54e-3 \pm 6.32e-5$ | $6.92e-3 \pm 1.31e-4$ |
| | | $Ns:Nt$ | $500:50$ | $500:50$ |
| Allen-Cahn | $128 \times 200$ | RL2E | $1.18e-3 \pm 5.47e-5$ | $1.60e-3 \pm 3.95e-5$ |
| | | $Ns:Nt$ | $1000:100$ | $1000:100$ |
| Maxwell's | $128 \times 100$ | RL2E | $1.05e-3 \pm 3.64e-5$ | $1.33e-2 \pm 3.26e-4$ |
| | | $Ns:Nt$ | $1000:100$ | $1000:100$ |
| Schrödinger | $128 \times 100$ | RL2E | $5.62e-4 \pm 4.44e-5$ | $3.82e-3 \pm 2.01e-4$ |
| | | $Ns:Nt$ | $1000:100$ | $1000:100$ |
| Navier-Stokes | $64^2 \times 100$ | RL2E | $9.50e-3 \pm 1.58e-4$ | $5.70e-3 \pm 2.71e-4$ |
| | | $Ns:Nt$ | $500:50$ | $500:50$ |
| Shallow Water | $64^2 \times 100$ | RL2E | $7.65e-3 \pm 2.13e-4$ | $1.61e-2 \pm 4.49e-4$ |
| | | $Ns:Nt$ | $500:50$ | $500:50$ |

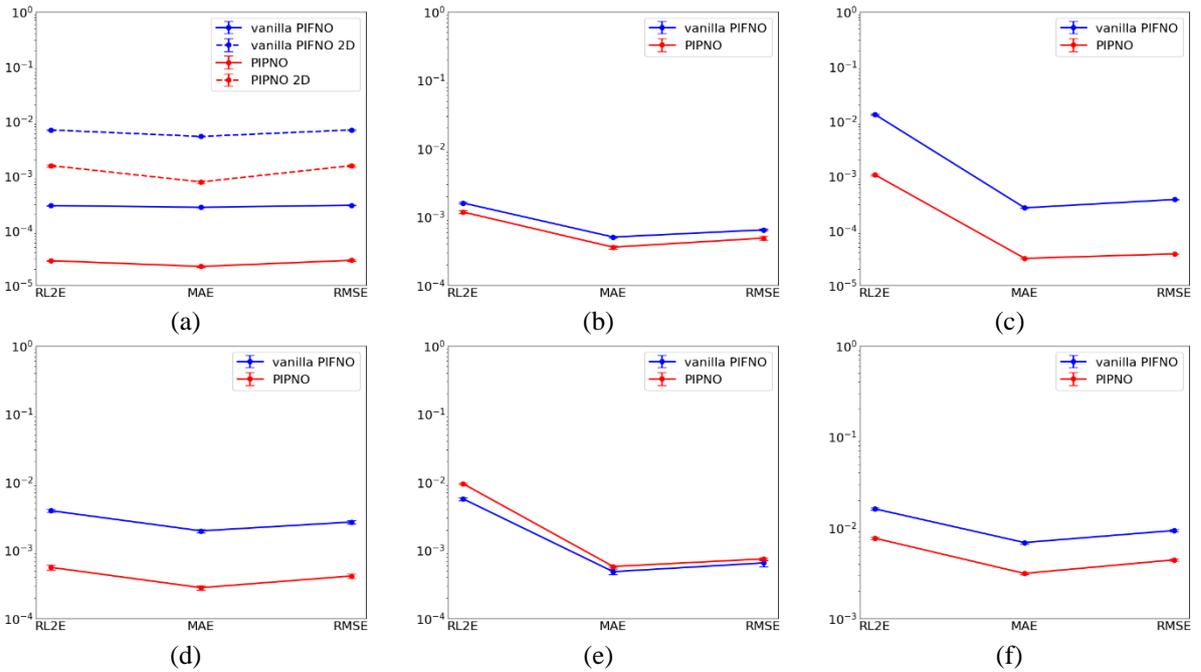

**Figure 4** The error plots of all the PDEs predictions on testing datasets and the comparisons between different physics-informed neural operators and reference solution for all the PDEs numerical examples. (a)

Consolidation equation; (b) Allen-Cahn equation; (c) Maxwell's equations; (d) Schrödinger equation; (e) Navier-Stokes equation; (f) Shallow Water equations.

### 4.1 Consolidation equation

For illustration, the Terzaghi's consolidation equation with varying initial conditions is considered as a first numerical example. The consolidation equation is a linear (when the equation coefficient is a constant) and parabolic PDE and is commonly regarded as one of the most fundamental equations in the field of geotechnical engineering. The equation describes an important theory for the consolidation behaviour of soils, which is used to study the deformation of soils under loading over time and the process of dissipation of pore water pressures. It provides a substantial theoretical basis for the consolidation process simulation, settlement prediction, and settlement control procedure design of diverse types of infrastructures in geotechnical engineering. Furthermore, the numerical simulations in this section also provide a prominent reference for heat transmission physics since the consolidation and heat transfer equations share the same equation form. The one-dimensional (1D) and two-dimensional (2D) consolidation equation with specified boundary conditions is expressed mathematically as follows:

$$\begin{cases} \frac{\partial u}{\partial t} = C_v \nabla^2 u, x \in \Omega, t \in [0,1] \\ u(x,0) = u_0(x), x \in \Omega \\ u(0,t) = u(1,t) = u_b(t), t \in [0,1] \end{cases} \tag{28}$$

Where the spatial domain is defined as $\Omega = [0,1]^d$, $d = 1,2$ for 1D and 2D scenarios, respectively. $\nabla^2$ is the Laplace operator. $u_0(x)$ is the initial condition and input into the physics-informed neural operators as a training variable, $u_b(t)$ denotes the periodic boundary condition, and the coefficient of this equation is set as $C_v = 0.01$. In the consolidation equation, $u$ represents the excess pore water pressure in the consolidating layer, which is commonly used to simulate soil consolidation and settlement, as well as one of the benchmarks to verify different computational algorithms. The ground truth dataset is generated employing the fourth-order Runge-Kutta FDM method with $dx = 1/128$ and $dt = 1 \times 10^{-4}$ in the whole spatiotemporal domain through the utilisation of code compilation platforms, such as Matlab or Python. For different initial conditions, the neural operator aims to learn the operator $G: u_0(x) \to u(x,t)$, where G is the mapping between given initial functions $u_0(x)$ and spatiotemporal solutions $u(x,t)$ at the exact time. To generate a number of initial conditions for use in the simulation, random samples are taken from a one-dimensional Gaussian Random Field (GRF) with a Matérn kernel. The kernel is converted to Fourier space in order to match the periodic boundary conditions. This choice of Gaussian random field is the same as that presented in the FNO paper(43, 53, 59). The generated initial conditions satisfy the following relation: $u_0(x) \sim k_{GRF}(x, \sigma, l, \nu)$, where the smoothness parameter $\nu$ dictates the differentiability, length scale $l$ controls the range of correlations, and variance $\sigma$ determines amplitude of the generated field and they are set as $(\sigma, l, \nu) =$ (0.5,0.1, ∞). This setting makes the Matérn kernel the same as the radial basis function kernel used in paper [34], defined as $k(x_1, x_2) = e^{\frac{-\|x_1-x_2\|^2}{2l^2}}$.

Thereafter, a total of 1100 GRF random samplings are conducted, with 1000 samples utilized as the training dataset and the remaining 100 samples employed as the testing dataset. During the training phase, the small subset of the 1000 training simulations is selected as a mini-batch, with a batch size of 20. Gradient descent optimisation is performed in each epoch. During the testing phase, each testing sample is evaluated individually, resulting in a test batch size of 1. Additionally, 1100 real ground truth simulations are conducted as a control basis for measuring model performance. The accuracy of the model are evaluated through physical reality while training with randomly generated instances of the initial conditions. Meanwhile, multiple repetitions of experiments are conducted to identify the optimal adjustable hyperparameters, which ensures consistency throughout the study. The mean square error (MSE) loss value is approximately $10^{-8}$ at the final epoch of the model training, which indicates that the discrepancy between the predicted and true physics is at a terrific level. This suggests that PIPNO exhibits a commendable performance in this problem. In this study, it is demonstrated that the size of the training dataset has a significant impact on the test results. In particular, prior studies have also confirmed that initial data with higher amplitude and more intense features are often challenging to process. Furthermore, it is also observed that augmenting the quantity of initial data leads to enhanced outcomes, notably in terms of the model's capacity for generalisation, test accuracy, and robustness. This can be attributed to the highly nonlinear nature of the neural network, as well as effective training. Although neural operators have been demonstrated to possess super-resolution imputation capabilities, the model employed in this study maintains a fixed resolution for both spatial and temporal discretisation. Furthermore, the numerical example illustrates that well-trained models can effectively capture the physical spatiotemporal evolutionary behaviour and demonstrate excellent performance in predicting the trajectories of real physical systems in the testing dataset.

Figure 5 depicts the visualisation results of the predictions for four representative testing samples of initial conditions in one-dimensional consolidation scenarios, with a spatiotemporal resolution of $128 \times 100$. It is evident that the predictions of the PIPNO exhibit a notable degree of agreement with the corresponding reference solutions. Specifically, it can be observed that the trained model is capable of rapid and precise inferences on previously unobserved initial conditions. Furthermore, all predictions align with the global pattern of the actual physical system. This indicates that PIPNO has effectively learned the operator that maps the function space from the initial function to the solution function, which describes the underlying physical laws. On the other hand, the absolute errors associated with these predictions demonstrate that neural operators can provide relatively low levels of error at most locations and times. Additionally, the results of the prediction accuracy are presented in Table 1 and Figure 4, which compares the PIPNO framework with the reference baseline. The average relative $L_2$ error and its standard deviation of the neural operator predictions on the training dataset are employed to quantify the overall prediction error of the model and assess its degree of fluctuation. The reference baseline PIFNO yielded an overall generalisation error $2.87e-4 \pm 5.01e-6$, whereas PIPNO achieved an overall generalisation error $2.82e-5 \pm 8.88e-7$. This provides compelling evidence of the enhanced performance and stability of the proposed method.

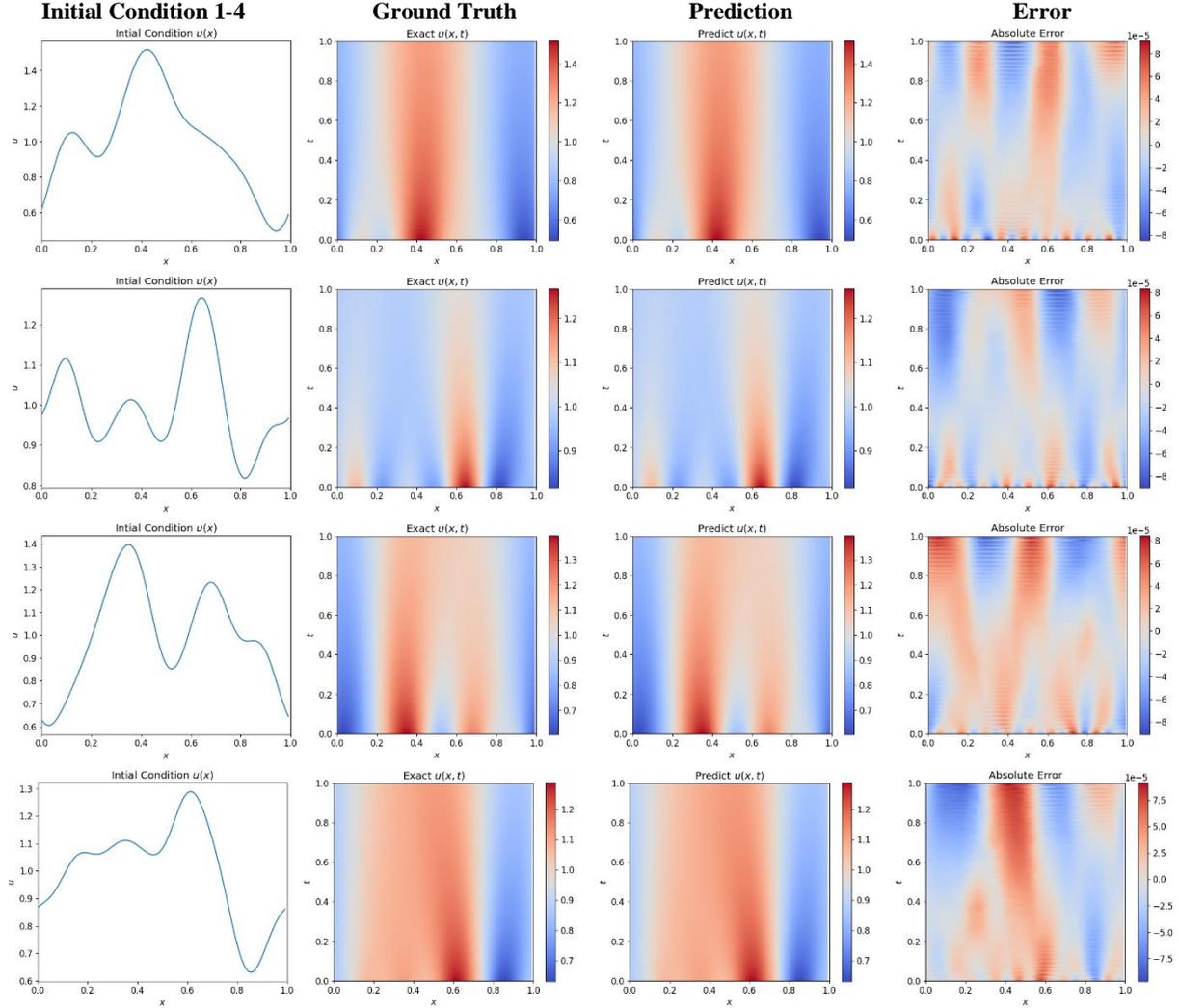

**Figure 5** The results of the 1D consolidation equation demonstrate the mapping from disparate initial conditions $u_0(x)$ to corresponding solutions $u(x,t)$ within a domain with a spatiotemporal resolution of 128 × 100. The presented data set includes the initial conditions, the corresponding ground truth, the PIPNO predictions, and error plots illustrating 4 distinct representative instances among the 100 testing samples.

In the two-dimensional scenario, the spatiotemporal resolution is $64 \times 64 \times 100$. Figure 6 presents the results comprehensively, showing the solution inference for four different initial conditions in the testing dataset. It can be observed that the discrepancies in the results are minimal and are predominantly concentrated around the peak of the function. As a consequence of the increase in data dimensionality, the training time of the neural operator is significantly longer than that observed in the one-dimensional case. Nevertheless, PIPNO continues to demonstrate excellent performance, with a mean relative $L_2$ error of $1.54e-3$ and a deviation of $6.32e-5$. In comparison, its counterpart exhibits a mean relative $L_2$ error of $6.92e-3$ and a deviation of $1.31e-4$. Two-dimensional problems are more general and representative in the study of physics problems. These findings suggest that neural operators have good operator learning ability in two-dimensional function spaces. This provides a promising avenue for capturing other two- or even high-dimensional dynamical systems and represents a robust complement and alternative to traditional numerical methods.

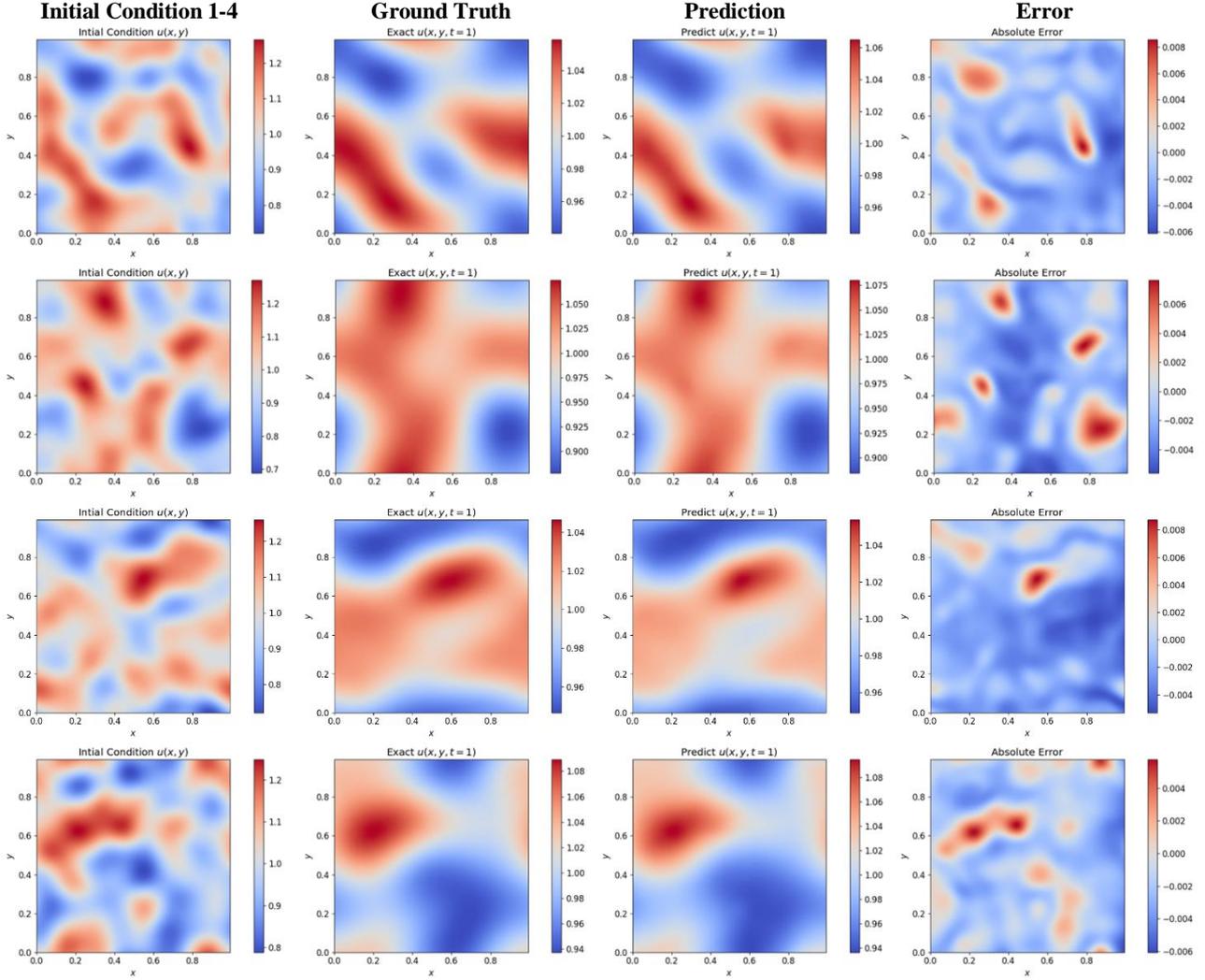

**Figure 6** The results of the 2D consolidation equation demonstrate the mapping from disparate initial conditions $u_0(x, y)$ to corresponding solutions $u(x, y, t)$ within a domain with a spatiotemporal resolution of 64 ×64 × 100. The presented data set includes the initial conditions, the corresponding ground truth, the PIPNO predictions, and error plots illustrating 4 distinct representative instances among the 50 testing samples.

### 4.2 Allen-Cahn equation

The Allen-Cahn equation is a parabolic nonlinear PDE that describes phase separation and interfacial dynamics in the phase-field model for studying phase transitions in materials science. It describes reaction-diffusion phenomenon and is commonly used to model a range of problems, including interface evolution (multicomponent alloys / chemical reactions / crystal growth), image segmentation, shape optimisation, etc. The Allen-Cahn equation and conditions are given below:

$$\begin{cases} \frac{\partial u}{\partial t} = \epsilon \frac{\partial^2 u}{\partial x^2} + u - u^3, x \in [0,1], t \in [0,2] \\ u(x,0) = u_0(x), x \in [0,1] \\ u(0,t) = u(1,t) = u_b(t), t \in [0,2] \end{cases} \tag{29}$$

Where the coefficient is set as $\epsilon = 0.0001$, $u_b(t)$ represents the periodic boundary condition. $u_0(x)$ is the initial condition generated by GRF, and its kernel takes the form when the smoothness parameter $v$ is infinite:

$$k(x) = \sigma^2 e^{\frac{-\|x_1 - x_2\|^2}{2l^2}} \tag{30}$$

Where $(\sigma, l) = (0.4, 0.1)$. This method actually remains consistent with the GRF for the first numerical example. To obtain the ground truth data for comparison, the previously generated random initial condition field is employed to evolve the PDE using fourth-order Runge-Kutta method along the temporal domain and central finite difference method in the spatial domain, thereby obtaining the ground truth solutions with high-order accuracy. The resolution for ground truth generation is set to $\delta x = 1/128$ and $\delta t = 1 \times 10^{-5}$. As with the preceding example, the objective is to identify an operator that maps representing the initial condition to the space-time solution at specific time point, namely $G: u_0(x) \to u(x,t)$. The simulation of $u_0(x)$ is conducted using the GRF. Subsequently, a finite difference-based script is employed to ascertain the ground truth while the neural operators are trained with a spatiotemporal resolution $128 \times 200$ using the initial conditions on the training dataset and making inferences on the testing dataset. The results are presented in Figure 7, while the comparative study is shown in Table 1 and Figure 4.

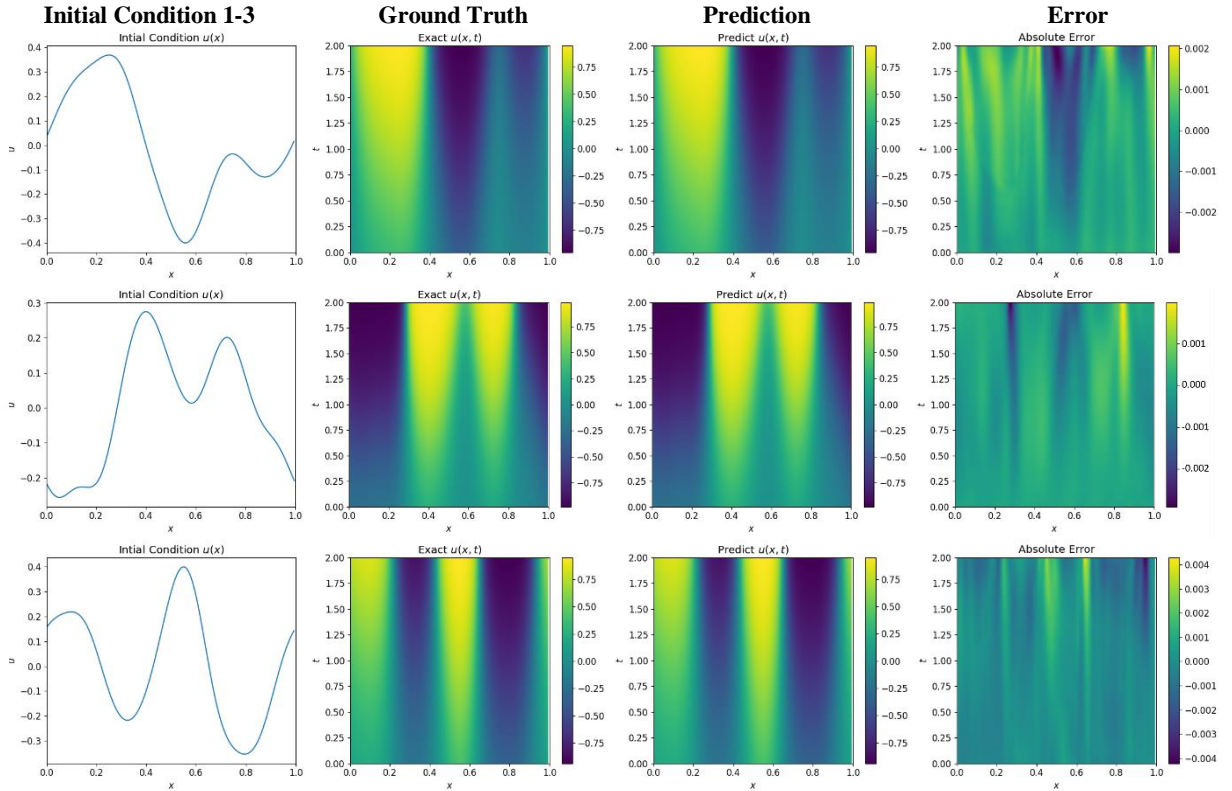

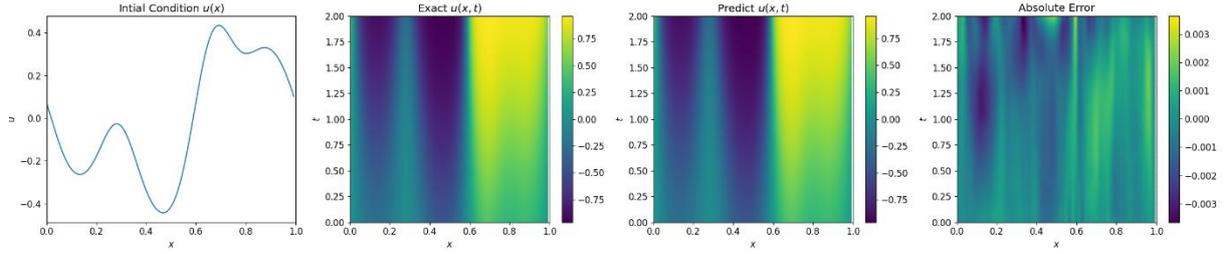

**Figure 7** The results of the 1D Allen-Cahn equation demonstrate the mapping from disparate initial conditions $u_0(x)$ to corresponding solutions $u(x,t)$ within a domain with a spatiotemporal resolution of 128 × 100. The presented data set includes the initial conditions, the corresponding ground truth, the PIPNO predictions, and error plots illustrating 4 distinct representative instances among the 100 testing samples.

The challenge in solving the Allen-Cahn equation primarily arises from the nonlinear reaction terms, sharp interfaces, and multiscale dynamics. When ϵ is exceedingly small, the interface becomes exceedingly thin, and the solution exhibits a significant variation in proximity to the interface, resulting in a sharp interface limit. Furthermore, the nonlinear reaction terms $u - u^3$ inherent to the equation enhance the intricacy of the solution, potentially impeding the convergence of the numerical algorithm. The results also demonstrate that the spatiotemporal evolutionary solutions of the equation exhibit distinct phase separation characteristics, with an increasing prevalence and prominence of multi-scale nonlinear features over time, such as zero and high gradients. Nevertheless, the neural operators continue to effectively capture these physics features. Figure 7 depicts the real and predicted solutions. The visual consistency between the predicted output and the ground truth, as well as the prediction error plots on the domains corresponding to the inferred realisations for the four different initial conditions, demonstrates the excellent performance of PIPNO for this problem. Table 1 and Figure 4 present the findings of the comparative study between physics-informed neural operators. The observations in the charts serve to reinforce the cognitive that PIPNO holds an excellent capacity to learn function mappings, all of these values are superior to those of the reference baseline. The mean relative $L_2$ error of PIPNO over the 100 testing samples is $1.18e - 3$, with a small standard deviation of only $5.47e - 5$, while the PIFNO receives a mean relative $L_2$ error of $1.60e - 3$ with a standard deviation of $3.95e - 5$. Physics-informed parallel neural operators get a better performance in learning and inferring dynamics.

### 4.3 Maxwell's equations

The Maxwell's equations constitute the fundamental theoretical framework for the description of electromagnetic fields, which consist of 4 PDEs. This set of equations describes a coupled nonlinear electromagnetic physical system, wherein the electric and magnetic fields are coupled to each other and evolve simultaneously in time and space, thereby contributing to the inherent complexity of comprehension. They are extensively employed for the derivation of the generation, variation, and interaction of electric and magnetic fields. Maxwell's equations form the foundation of electromagnetism and are a vital component of modern physics. They provide a crucial theoretical basis for various fields, including power engineering, electronics, computer technology, astronomy, medicine, and many other scientific and engineering disciplines. The general form of Maxwell's equations and their corresponding conditions in this section can be expressed as follows:

$$\begin{cases} \nabla \cdot (\epsilon E) = \rho_f; \nabla \cdot (\mu H) = 0 \\ \nabla \times E = -\mu \frac{\partial H}{\partial t}; \nabla \times H = J_f + \epsilon \frac{\partial E}{\partial t} \\ E(x,0) = E_0(x); H(x,0) = 0 \\ E(0,t) = E(1,t) = E_b(t); H(0,t) = H(1,t) = H_b(t) \\ x \in [0,1], t \in [0,1] \end{cases} \quad (31)$$

Where $E$ represents the electric field intensity, $H$ denotes the magnetic field intensity, $\rho_f = 0$ is the charge density per unit volume, $\epsilon = 1$ is the dielectric constant, $\mu = 1$ represents the magnetic permeability, and $J_f = 0$ denotes the current density source term. The mathematical operators $\nabla \cdot$ and $\nabla \times$ represent the divergence and curl of the corresponding physical field, respectively. $E_b(t)$ and $H_b(t)$ denote specific boundary conditions for the electric and magnetic fields, respectively. The settings outlined above are designed to streamline the simulation and computation processes of the model. In Equation 41, the first line represents Gauss's law for the electric and magnetic fields, respectively, and the second line is Faraday's law of electromagnetic induction and Ampere-Maxwell's law, respectively. In this section, a one-dimensional cavity model filled with a homogeneous medium is selected to validate the effectiveness and accuracy of the proposed deep learning approach for time-domain electromagnetic field simulation. Based on the above settings, there is no divergence of the electric and magnetic fields as the electromagnetic fields propagate through the medium, and Gauss's law becomes an implicit condition that is always satisfied, rather than a dynamic equation, and is often not computed explicitly. To maintain generality, the electric field $E_y$ and the magnetic field $H_z$ are focused as the primary variables in this study. This means only these components, which are functions of the $z$ direction and time $t$, are considered. Maxwell's equations, in this particular case, can be simplified for these specific components, resulting in a straightforward model. Therefore, in one-dimensional scenarios, the curl of the electric field is simplified to $\nabla \times E = \frac{\partial E_y}{\partial x}$, while the curl of the magnetic field is simplified to $\nabla \times H = -\frac{\partial H_z}{\partial x}$, and replace them into Equation (31) as the final electromagnetic field computational model. For the sake of convenience, the notation $E$ and $H$ will be used consistently throughout this section.

In this example, the objective is to learn function mapping operators. The Maxwell's equations consist of a set of coupled PDEs. In other words, the operator is the mapping from an initial function to two solutions $G: E_0(x) \to [E(x,t), H(x,t)]$. Learning such coupled operators is often more challenging than that in uncoupled physics due to the complex interactions of coupled physical fields. $E_0(x) \sim k_{GRF}(x, \sigma, l, \nu)$ is the initial condition generated by GRF that is the same with the former sections, where $(\sigma, l, \nu) = (0.1, 0.1, \infty)$, which is used as the input into the physics-informed neural operators as a training variable. In addition, to obtain the ground truth data for comparison, the reference solution is generated by fourth-order Runge-Kutta FDM with $dx = 1/128$ and $dt = 1 \times 10^{-5}$ in the whole defined spatiotemporal domain.

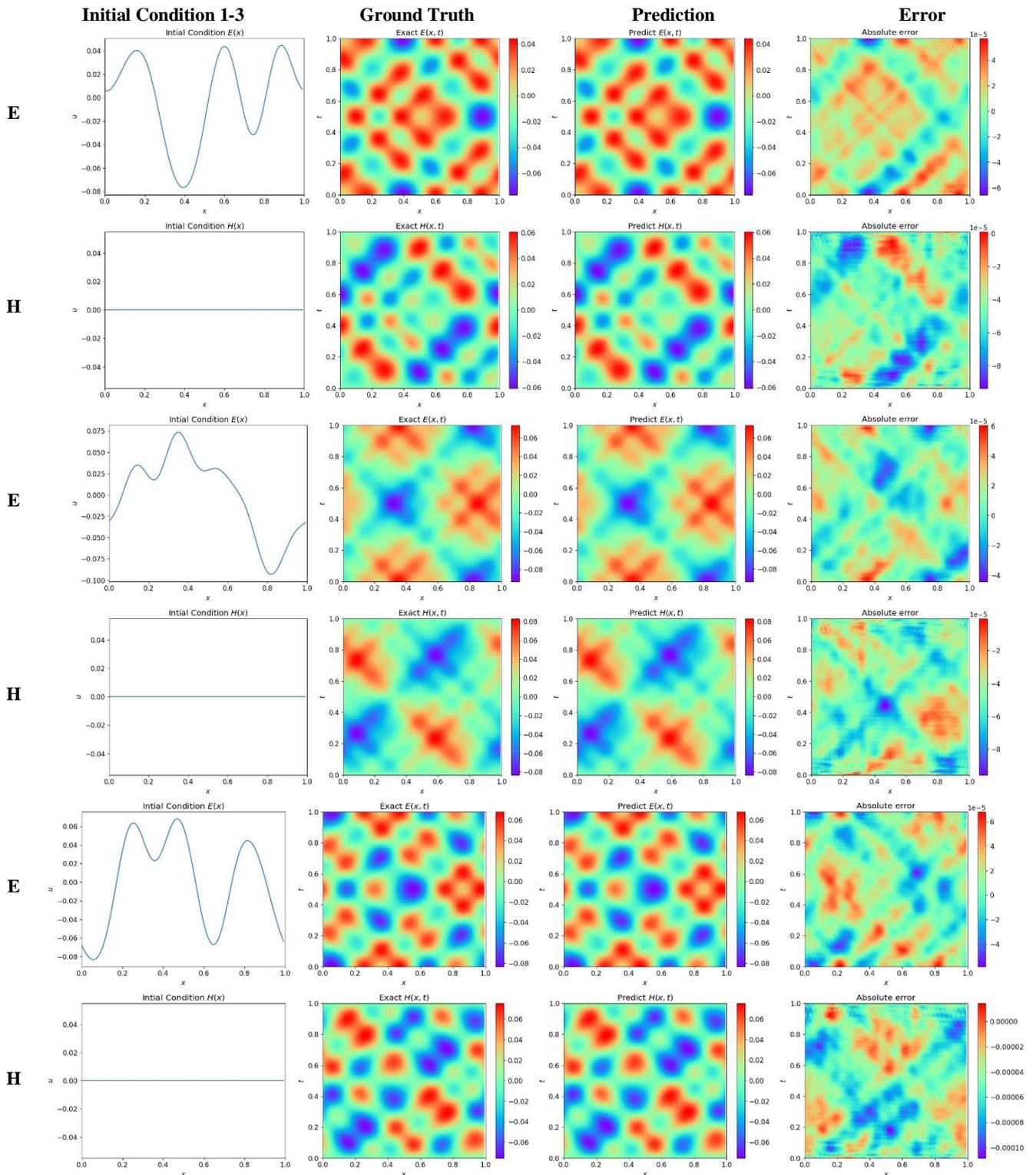

**Figure 8** The results of the 1D Maxwell's equations demonstrate the mapping from disparate initial conditions $E_0(x)$ to corresponding solutions $[E(x,t), H(x,t)]$ within a domain with a spatiotemporal resolution of 128 × 100. The presented data set includes the initial conditions, the corresponding ground truth, the PIPNO predictions, and error plots illustrating 3 distinct representative instances among the 100 testing samples.

The difficulty in solving Maxwell's equations is mainly because it is a highly coupled system of vectorial PDEs involving spatiotemporal interactions of electric and magnetic fields with multiscale characteristics. The solutions of equations are usually prone to significant high-frequency features, especially when describing electromagnetic field propagation and oscillatory phenomena, which can also be easily observed in the results. In this study, the spatiotemporal resolution of neural operators for this study is $128 \times 100$. The demonstration results presented in Figure 8 include the input function of initial conditions, the ground truth solution, the neural operator predictions, and the prediction errors. These results clearly illustrate the effectiveness of the proposed framework in learning coupled PDEs operators. In all 4 different testing cases, the predictions made by the framework show a very strong agreement with the corresponding reference solutions as the absolute errors are relatively low. Moreover, from the results shown in Table 1, the average relative $L_2$ error of PIPNO over the 100 testing samples is $1.05e-3$, with a small standard deviation of only $3.64e-5$, while PIFNO produces a higher average relative $L_2$ error $1.33e-2 \pm 3.26e-4$. Therefore, Physics-informed parallel neural operators get a better performance in learning and inferring dynamics, making them a good surrogate model according to the results comparisons.

### 4.4 Schrödinger equation

The Schrödinger equation serves as the fundamental basis for quantum mechanics. It captures the wave-like behavior of elementary particles and attributes their behavior to the propagation and superposition of wave functions. Additionally, the equation describes how the wave functions of particles evolve. While it is akin to Newton's second law in classical mechanics, the Schrödinger equation is specifically designed to describe the wave functions of quantum systems. This essential equation has a significant impact across various fields, including physics, chemistry, materials science, and computer science, making it a crucial pillar of modern physics. The time-dependent Schrödinger equation and its conditions considered in this study are given below. Certain simplifying assumptions are made to the equations for the convenience of numerical simulations.

$$\begin{cases} i\hbar \frac{\partial \psi}{\partial t} = \widehat{H}\psi, x \in [0,1], t \in [0,1] \\ \psi(x,0) = \psi_0(x), x \in [0,1] \\ \psi(0,t) = \psi(1,t) = \psi_b(t), t \in [0,1] \end{cases} \quad (32)$$

Where $\psi(x,t)$ represents the wave function, which describes the probability amplitude of a particle at spatial position $x$ and time $t$; $i$ is the imaginary unit; $\hbar = 1$ is the approximate Planck's constant; and $\widehat{H} = -\frac{\hbar^2}{2m}\nabla^2 + V(x,t)$ stands for the Hamiltonian operator, which represents the total energy of the system, $m$ is set as 1, $V(x,t) = -|\psi(x,t)|^2$ represents the potential field in the system. $u_0(x)$ is the initial condition, and $\psi_b(t)$ represents the specific boundary condition. Certainly, if the real and imaginary parts of $\psi$ are denoted by $u$ and $v$, respectively, namely $\psi = u + iv$ and $|\psi|^2 = u^2 + v^2$, then $u$ and $v$ satisfy a coupled relationship in the original governing equation. To create various initial conditions for the simulation, random samples are drawn from a one-dimensional Gaussian Random Field (GRF) using a Matérn kernel. The generated initial conditions satisfy the following relation: $u_0(x) \sim k_{GRF}(x, \sigma, l, \nu)$, where $(\sigma, l, \nu) = (0.5, 0.4, \infty)$. The ground truth dataset is generated using the fourth-order Runge-Kutta finite difference method, with $\delta x = 1/128$ and $\delta t = 1 \times 10^{-5}$ defined across the entire spatiotemporal domain. The neural operator is designed to learn the mapping operator $G: \psi_0(x) \to [u(x,t), v(x,t)] = \psi(x,t)$, which relates given initial functions to their corresponding spatiotemporal solutions at specific times.

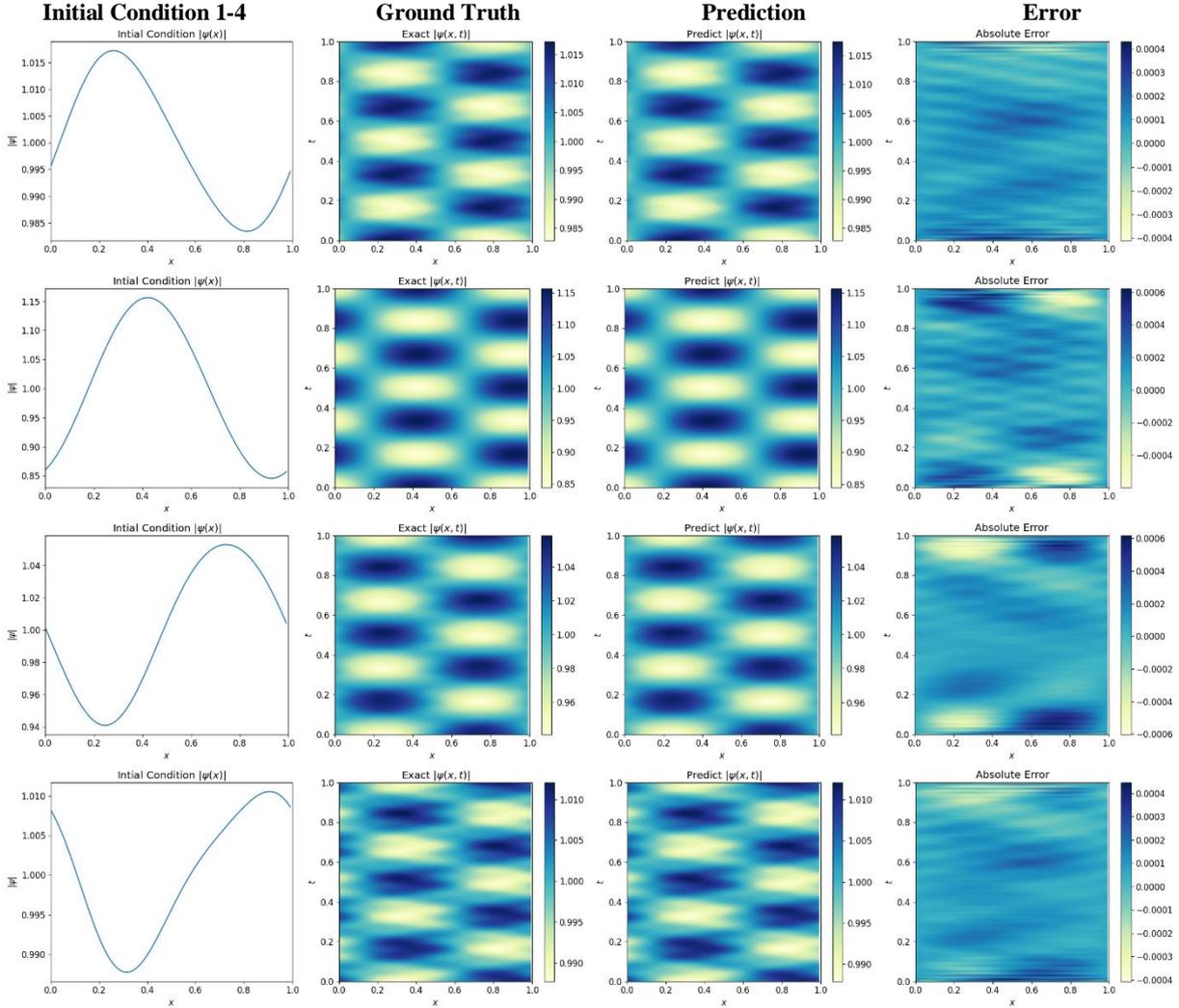

**Figure 9** The results of the 1D Schrödinger equation demonstrate the mapping from disparate initial conditions $\psi_0(x)$ to corresponding solutions $\psi(x,t)$ within a domain with a spatiotemporal resolution of 128 × 100. The presented data set includes the initial conditions, the corresponding ground truth, the PIPNO predictions, and error plots illustrating 4 distinct representative instances among the 100 testing samples.

The difficulty in solving the time-dependent Schrödinger equation primarily arises from its highly nonlinear and strongly coupled characteristics, especially in high-dimensional spaces or many-particle systems, where the complexity of the wave function makes the computational resources extremely demanding. The wave function is sometimes characterised by high-frequency oscillations, which are also reflected in the presented results, and the time evolution process needs to satisfy both accuracy and stability, while potentials with time-dependent or non-smooth conditions increase the difficulty of the numerical solution. High-resolution grids and refined time steps are also required to capture the fine features of the quantum states, which may further increase the computational burden. In addition, it is also found from the numerical experiments that the choice of initial conditions can have a significant impact on the accuracy and results, with non-smooth fluctuating initial conditions being more inclined to produce high-frequency oscillatory features. Thus, relatively smooth initial conditions are chosen for demonstration in this numerical case. Figure 9 illustrates the visualisation results of the predictions for four representative testing samples of initial conditions, with a spatiotemporal resolution of $128 \times 100$. The predictions of the PIPNO show a strong agreement with the corresponding reference solutions. Notably, the trained model demonstrates the ability to make quick and accurate inferences based on previously unobserved initial conditions. Additionally, all predictions are consistent with the global pattern of the actual physical system. This suggests that PIPNO has successfully learned the operator that maps the function space from the initial function to the solution function, effectively capturing the underlying physical laws. Moreover, the absolute errors associated with these predictions indicate that neural operators can maintain relatively low error levels across most locations and times. The prediction accuracy results are summarized in Table 1 and illustrated in Figure 4, which compares the PIPNO framework with the reference baseline. To quantify the overall prediction error of the model and assess its variability, the average relative error and its standard deviation are also reported based on the neural operator predictions from the training dataset. The reference baseline PIFNO yielded an overall generalisation error $3.82e-3 \pm 2.01e-4$, whereas PIPNO achieved an overall generalisation error $5.62e-4 \pm 4.44e-5$. The findings clearly demonstrate the improved performance and stability of the proposed method.

### 4.5 Navier-Stokes equation

The Navier-Stokes equations are a set of second-order nonlinear parabolic PDEs that form the foundation of fluid dynamics. These equations describe the motion of fluids and are employed to model their time-dependent behavior. They find applications across a wide range of scientific and engineering disciplines, including meteorology, oceanography, mechanical engineering, aerospace engineering, environmental engineering, biology, and medical engineering, among others. The Navier-Stokes equations are crucial for studying turbulence, vortex dynamics, and nonlinear phenomena. Specifically, the two-dimensional unsteady Navier-Stokes equations for incompressible fluids are utilized to characterize the physical dynamic evolution of these fluids. This system consists of a mass conservation equation and a momentum conservation equation, adhering to the incompressibility condition. Its vorticity-velocity formulation and conditions are represented by the following equations:

$$\begin{cases} \dfrac{\partial u}{\partial x} + \dfrac{\partial v}{\partial y} = 0 \\ \dfrac{\partial \omega}{\partial t} + u\dfrac{\partial \omega}{\partial x} + v\dfrac{\partial \omega}{\partial y} = \vartheta\left(\dfrac{\partial^2 \omega}{\partial x^2} + \dfrac{\partial^2 \omega}{\partial y^2}\right) + f(x) \\ \omega(x, y, 0) = \omega_0(x, y) \\ \omega(0, y, t) = \omega(1, y, t) = \omega_b^x(y, t); \omega(x, 0, t) = \omega(x, 1, t) = \omega_b^y(x, t) \\ x, y \in [0,1], t \in [0,1] \end{cases} \quad (33)$$

Where $u$ and $v$ denote the vector components of the fluid velocity field in the $x$ and $y$ direction, respectively. $\omega$ represents the vorticity field of the fluid, $\vartheta = 0.05$ is the kinematic viscosity coefficient of the fluid, and $f(x) = 0$ corresponds to the external source term function. $\omega_b^x(y, t)$ and $\omega_b^y(x, t)$ are the given boundary conditions of fluid vorticity in the $x$ and $y$ direction, respectively. The vorticity-velocity equation exhibits a nonlinear coupling property, where the vorticity field and velocity field are interconnected by $\omega = \nabla \times \boldsymbol{u}$, where $\nabla \times$ is the curl operator and $\boldsymbol{u} = (u, v)$ is the velocity field. To figure out the time evolution of vorticity, the stream function $\psi$ is first constructed, which in conjunction with the vorticity satisfies the following Poisson equation:

$$\nabla^2 \psi = -\omega \quad (34)$$

Where $\nabla^2$ is the Laplace operator. Finally, the velocity field is solved from the relationship between the velocity and the stream function:

$$\boldsymbol{u} = \nabla \times \psi \quad (35)$$

and then they are brought into the momentum conservation equation for vorticity. $\omega_0(x, y)$ is the initial condition generated by GRF, which is utilized as an input variable for the physics-informed neural operators during training. Specifically, the generated initial conditions satisfy the following relation: $\omega_0(x, y) \sim k_{GRF}(x, y, \sigma, l, \nu)$, where $(\sigma, l, \nu) = (0.6, 0.2, \infty)$. To obtain the ground truth data for comparison, the previously generated random initial condition field is used to evolve the PDE using the Fourier spectral method with forward Euler inference along the temporal domain. The resolution for ground truth generation is set to $\delta x = \delta y = 1/64$ and $\delta t = 1 \times 10^{-4}$. Similar to the previous example, the objective is to identify an operator that maps representing the initial condition to the space-time solution at a specific time point, namely $G: \omega_0(x, y) \to \omega(x, y, t)$. Subsequently, the neural operators are trained with a spatiotemporal resolution $64 \times 64 \times 100$ using the initial conditions on the training dataset and making predictions on the testing dataset. The results are presented in Figure *10*, while the comparative study is shown in Table 1 and Figure 4.

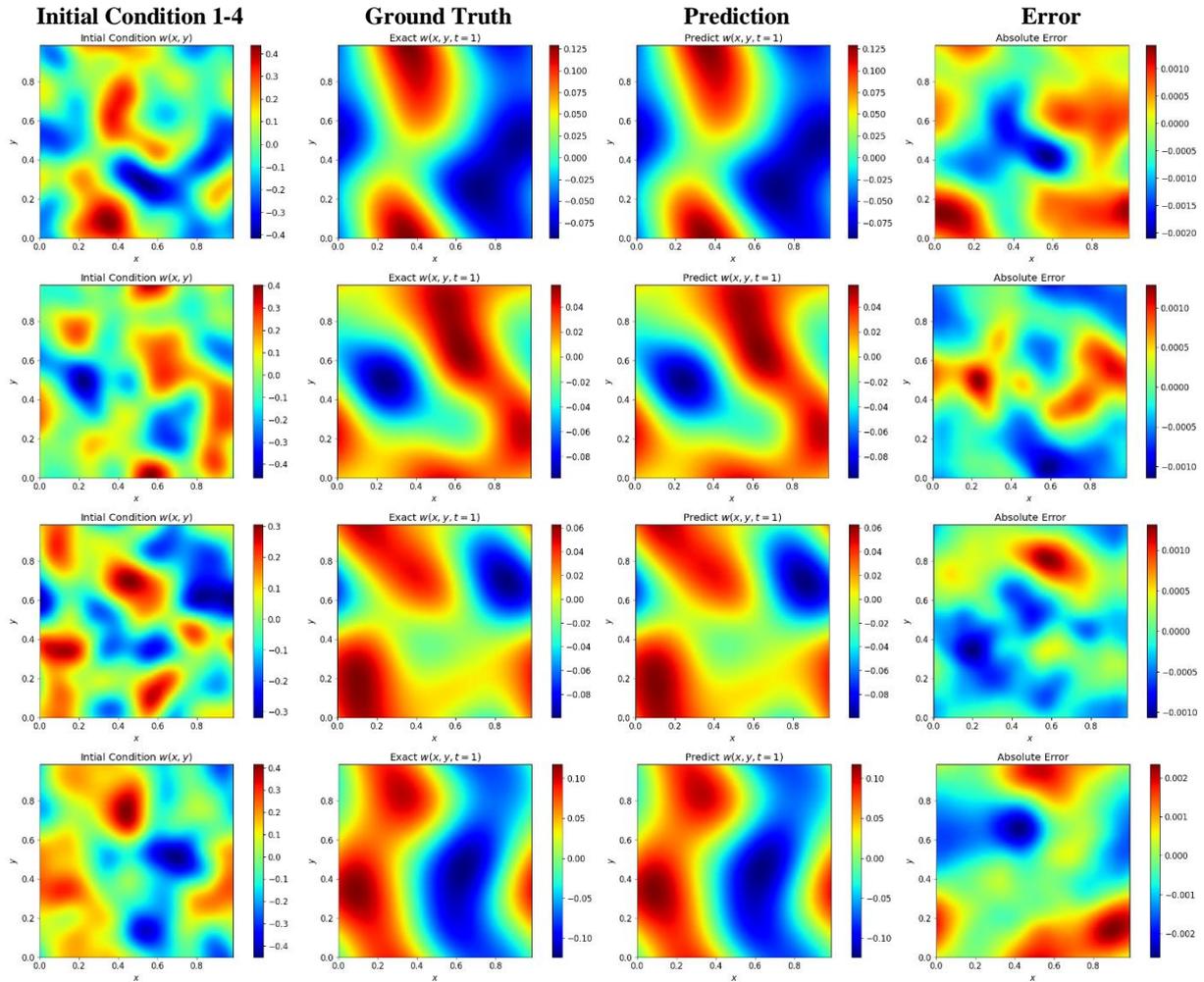

**Figure 10** The results of the 2D Navier-Stokes equation demonstrate the mapping from disparate initial conditions $\omega_0(x,y)$ to corresponding solutions $\omega(x,y,t)$ within a domain with a spatiotemporal resolution of 64 ×64 × 100. The presented data set includes the initial conditions, the corresponding ground truth, the PIPNO predictions, and error plots illustrating 4 distinct representative instances among the 50 testing samples.

The two-dimensional unsteady Navier-Stokes equations are a powerful tool for modelling fluid motion. However, their nonlinear and coupled nature makes finding analytical solutions a challenging computational task. As a result, numerical methods such as the finite difference method, finite volume method, and finite element method are commonly used to approximate solutions in practical applications. These traditional numerical methods often require complex algorithms and significant computational resources for implementation. Neural operators have recently emerged as a powerful alternative model, providing innovative solutions to traditional scientific challenges in the control of PDEs. Figure 10 offers the true and predicted solutions. For a given unseen scenario, the visual consistency between the predicted output and the ground truth, along with the prediction error plots for the domains corresponding to the inferred realisations under four different initial conditions, demonstrates the excellent performance of PIPNO for this problem. Table 1 and Figure 4 illustrate the findings of the comparative study on physics-informed neural operators. The observations in these charts support the idea that PIPNO has a strong ability to learn function mappings, with all values exceeding those of the reference baseline. Furthermore, from the prediction errors listed in the table, it can be observed that the average relative $L_2$ error of PIPNO over 50 testing samples is only about $9.50e-3 \pm 1.58e-4$, which is a much larger value compared to the previous examples. Furthermore, in this example, the PIFNO produces an average relative $L_2$ error of $5.70e-3 \pm 2.71e-4$ over 50 testing samples which is remarkably lower than that of PIPNO. The reason for this result lies in using the spectral method and decoupling the velocity in Fourier space to generate the reference solution, which imparts a frequency domain character to the data. It aligns well with the process of solving the PDE residuals in PIFNO. In other words, the principle of PIFNO is perfectly adapted to the spectral method, having a decoupling mechanism that resembles traditional methods. This consistency enhances the compatibility of PIFNO with the mathematical and physical structure of this problem.

### 4.6 Shallow Water equations

As the final numerical example, the two-dimensional nonlinear shallow water equations are considered. They are the basic set of equations describing the hydrodynamics of shallow water, which are suitable for situations where the depth of the fluid is much smaller than the horizontal scale, such as the flow of water in the near-shore region of rivers, lakes, and oceans. Therefore, they are commonly used to predict flood waves, simulate tides, tsunamis, and storm surges, study river scour, and simulate urban drainage systems. The two-dimensional nonlinear shallow water equations are derived from the conservation of mass and momentum in hydrodynamics, in which the water depth and velocity are tightly coupled in the mass and momentum conservation equations. As a core model in hydrodynamics, it can capture the macroscopic flow behaviour as well as reveal the microscopic fluctuation mechanism, providing a theoretical foundation and simulation tool for fluid mechanics, engineering applications, and environmental protection. Finally, the equations and conditions are described as below:

$$\begin{cases} \dfrac{\partial(\eta)}{\partial t} + \dfrac{\partial(\eta u)}{\partial x} + \dfrac{\partial(\eta v)}{\partial y} = 0 \\ \dfrac{\partial(\eta u)}{\partial t} + \dfrac{\partial}{\partial x}\left(\eta u^2 + \dfrac{1}{2}g\eta^2\right) + \dfrac{\partial(\eta uv)}{\partial y} = \vartheta\left(\dfrac{\partial^2 u}{\partial x^2} + \dfrac{\partial^2 u}{\partial y^2}\right) \\ \dfrac{\partial(\eta v)}{\partial t} + \dfrac{\partial(\eta uv)}{\partial x} + \dfrac{\partial}{\partial y}\left(\eta v^2 + \dfrac{1}{2}g\eta^2\right) = \vartheta\left(\dfrac{\partial^2 v}{\partial x^2} + \dfrac{\partial^2 v}{\partial y^2}\right) \\ \eta(x,y,0) = \eta_0(x,y), u(x,y,0) = 0, v(x,y,0) = 0 \\ \eta(0,y,t) = \eta(1,y,t) = \eta_b^x(y,t); \eta(x,0,t) = \eta(x,1,t) = \eta_b^y(x,t) \\ u(0,y,t) = u(1,y,t) = u_b^x(y,t); u(x,0,t) = u(x,1,t) = u_b^y(x,t) \\ v(0,y,t) = v(1,y,t) = v_b^x(y,t); v(x,0,t) = v(x,1,t) = v_b^y(x,t) \\ x,y \in [0,1], t \in [0,1] \end{cases} \quad (36)$$

Where $\eta$ represents the height of the water, $u$ and $v$ denote the vector components of the water velocity field in the $x$ and $y$ direction, respectively. $\omega$ represents the vorticity field of the fluid, $g = 1$ is the gravitational acceleration, and $\vartheta = 0.002$ is the kinematic viscosity coefficient of the water. $\eta_b^x(y,t)$ and $\eta_b^y(x,t)$, $u_b^x(y,t)$ and $u_b^y(x,t)$, $v_b^x(y,t)$ and $v_b^y(x,t)$ are the specific boundary conditions of the water height, water horizontal and vertical velocity in the $x$ and $y$ direction, respectively. In this study, the objective is to learn function mapping operators. The shallow water equations comprise a set of coupled PDEs. In this context, the operator is the mapping from an initial function to two resulting solutions $G: \eta_0(x,y) \rightarrow [\eta(x,y,t), u(x,y,t), v(x,y,t)]$. That is to say, the goal is to learn the function mapping operators that relate the height of the water, the horizontal velocity of the water, and the vertical velocity of the water through physical constraints and effective training. However, deriving the mapping of these three interrelated sub-functions from a single initial function is more complicated than the previously mentioned examples due to the nonlinear coupling. Learning coupled operators is often more challenging than uncoupled physics because of the complex interactions between coupled physical fields. $\eta_0(x,y) \sim k_{GRF}(x,y,\sigma,l,\nu)$ is the initial condition generated by GRF which is consistent with the previous sections, where $(\sigma, l, \nu) = (0.2, 0.1, \infty)$. The GRF-generated initial conditions are used as the input data into the physics-informed neural operators as a training variable. Furthermore, to acquire the ground truth data for comparison, the reference solution is obtained using the fourth-order Runge-Kutta FDM with $dx = dy = 1/64$ and $dt = 1 \times 10^{-4}$ in the whole defined spatiotemporal domain.

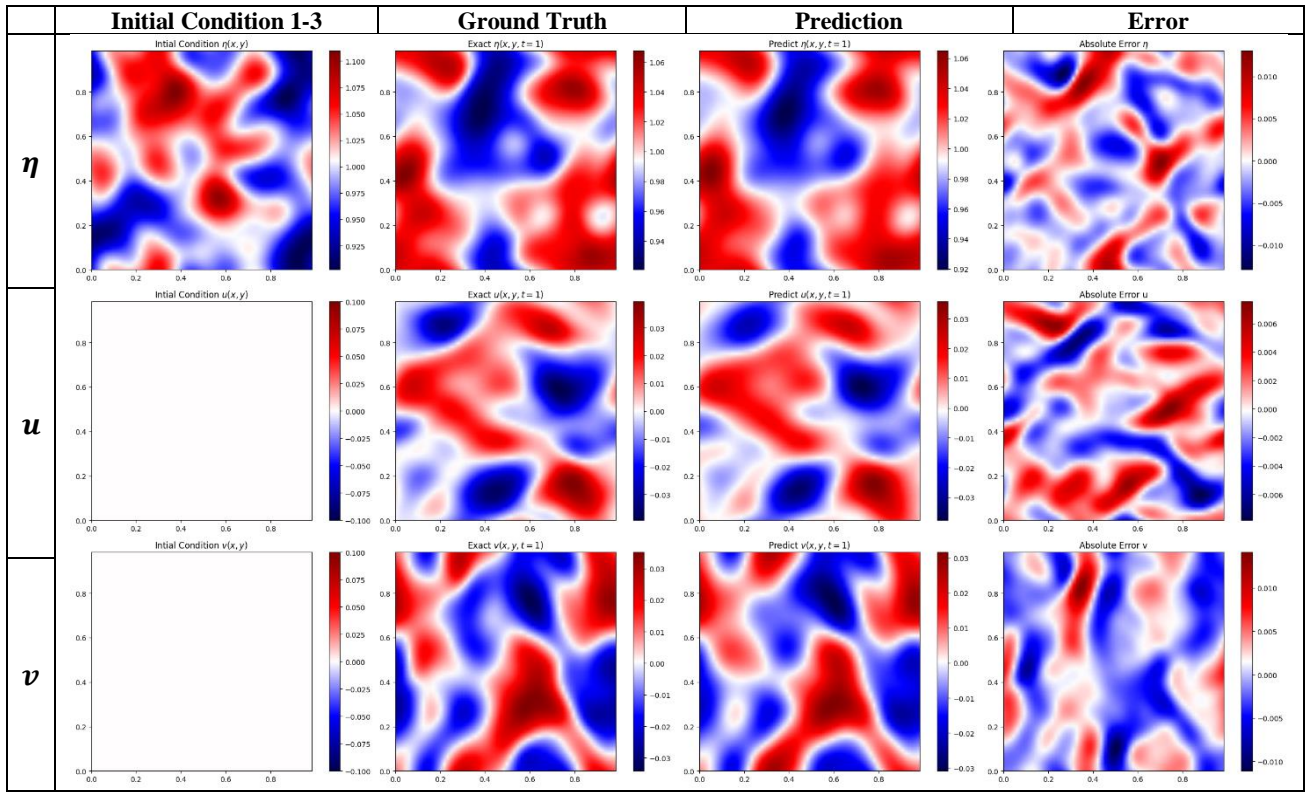

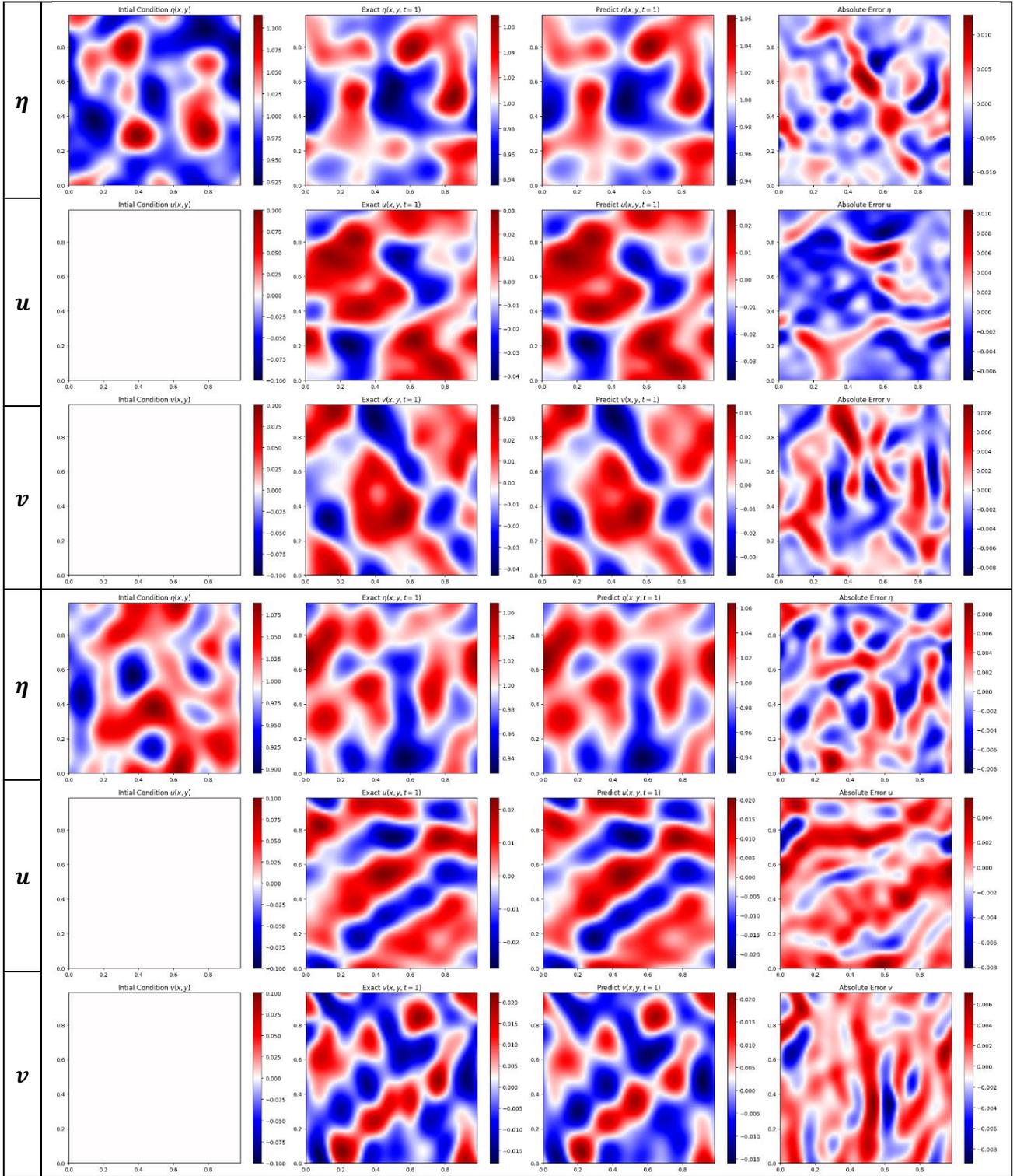

**Figure 11** The results of the 2D Shallow Water equation demonstrate the mapping from disparate initial conditions $\eta_0(x,y)$ to corresponding solutions $[\eta(x,y,t), u(x,y,t), v(x,y,t)]$ within a domain with a spatiotemporal resolution of 64 × 64 ×100. The presented data set includes the initial conditions, the corresponding ground truth, the PIPNO predictions, and error plots illustrating 3 distinct representative instances among the 50 testing samples.

The two-dimensional shallow-water equations are a set of nonlinear hyperbolic PDEs that involve multiple variables tightly coupled through convective terms and continuity equations. This strong coupling among the variables means they are closely related, leading to an entire system of equations that must be solved simultaneously. The coupling is also a significant challenge for numerical solutions, as it requires methods that can handle complex nonlinear interactions while maintaining the physical consistency of the solution. For results, the spatiotemporal resolution of neural operators for this study is $64 \times 64 \times 100$. The demonstration results shown in Figure 11 include the initial conditions, the ground truth solution, the neural operator predictions, and the prediction errors. These results effectively illustrate the capability of the proposed framework in learning coupled PDE operators. In all 4 testing cases, the predictions made by the framework exhibit strong agreement with the corresponding reference solutions, as the absolute errors are relatively low. Moreover, from the results shown in Table 1, the mean relative $L_2$ error of PIPNO over 50 testing samples is $7.65e-3$, with a small standard deviation of only $2.13e-4$, while PIFNO produces a higher average relative $L_2$ error $1.61e-2 \pm 4.49e-4$. This illustrates how effective our proposed neural operator is in addressing similar problems.

## 5. Conclusions

In this paper, the physics-informed parallel neural operator (PIPNO) under unsupervised learning in multiphysics systems are proposed as an innovative approach to operator learning. Previous studies have shown that neural networks can approximate any continuous nonlinear operator and learn function mappings in infinite-dimensional function spaces. PIPNO can make reasonable and accurate function mapping predictions for function solutions of PDEs given initial functions without any assumptions and prior knowledge. It can learn directly from control physics instead of being driven by ground truth data. Meanwhile, it is worth noting that PIPNO differs from the originally proposed PINN in that it can only solve a specific problem, which means that any change in the initial conditions, boundary conditions, geometry, or material characteristics requires solving the problem over again. Traditional PDE solvers, such as Finite Difference Methods and Finite Element Methods, similarly require repeated independent runs for different input conditions. However, PIPNO can learn a series of mappings for the family of PDEs. As a result, it can quickly obtain the target solution even if these conditions change, which makes it one of the most promising developments in computational mechanics. In addition, the original concept of operator learning was largely data-driven, making it highly suitable for problems involving large amounts of data. Currently, most contemporary research on operator learning still requires enormous amounts of data, even for physics-informed operator learning, which relies heavily on traditional PDE solvers for training. However, the reliance on substantial amounts of training data can be challenging, especially given the prohibitive cost or even unavailability of data generation, which also greatly limits its performance and development. In contrast, PIPNO allows access to more general and universal models. Theoretically, to obtain function mappings in infinite-dimensional Banach space and Hilbert space, PIPNO performs a series of updates to the inputs and then transforms them in different function spaces respectively to achieve learning under parallel function spaces. The parameterised integral kernel operators combined with nonlinear activation functions facilitate these operations, while the parameterised multilayer perceptrons are designed as the local fully connected neural networks. Essentially, this approach enables unsupervised learning without data support and improves performance in terms of accuracy and generalisation robustness, which allows PIPNO to outperform state-of-the-art operator frameworks in learning highly nonlinear operators for dynamics capture, by analysing the physical system defined by the governing equations. The PDEs, initial conditions and boundary conditions of the control physics system are the only exactly known preconditions and do not require any labelled data or data-driven learning. Such methods can be used as numerical solvers and surrogate models for parametric PDEs, which only need to be given precise physical information and have the potential to solve the corresponding inverse problems and to be the cornerstone of the next generation of large artificial intelligence models for computational science.

The results of the numerical experiments confirm the effectiveness of PIPNO in modelling physical systems. A series of examples covering multiple disciplines, such as geotechnical engineering, materials science, electromagnetism, quantum mechanics, fluid dynamics, etc., are studied using several PDEs. These equations include the Consolidation equation, the Allen-Cahn equation, the Maxwell's equation, the Schrödinger equation, the Navier-Stokes equation, and the Shallow Water equation. Many of them address highly nonlinear and strongly coupled physical problems, which are used to validate the operator learning capability of the proposed method in complex scenarios with multiple function mappings. Specifically, the initial conditions have remarkable influence on the prediction accuracy, especially in strongly nonlinear or uncoupled equations, smoother initial conditions lead to higher accuracy. The results from the numerical experiments align well with the actual physical dynamics, demonstrating the excellent performance and broad potential of PIPNO as a reliable surrogate model for simulating physical systems. Additionally, the real-time efficiency of the method makes it suitable for governing PDEs in time-dependent physical scenarios. PIPNO works across a wide range of applications and can be extended to fields such as physics-enhanced learning, digital twins, and the development of large-scale AI models for computational physics. Furthermore, this study illustrates that better kernel integration algorithms can be developed based on their complementary strengths to achieve efficient operator approximation while addressing different challenges and difficulties. However, despite its promising performance, PIPNO does have some notable drawbacks. Firstly, there is a lack of standardized procedures for determining the optimal weights of the loss function and for setting the most appropriate configuration parameters for the model. Secondly, numerical discretization in differentiations introduces inherent errors, and uniform grids may limit its applicability. Finally, the numerical experiments, such as in Navier-Stokes equation, reveal that the proposed method still struggles and performs poorly in some scenarios with more highly nonlinear and strongly coupled multiphysics systems, as it lacks a decoupling mechanism similar to those found in traditional methods. Consequently, the challenges mentioned above can be viewed as important goals for further investigation. By addressing these limitations, we can develop a powerful framework applicable to a variety of physical systems in science and engineering.

## CRediT authorship contribution statement

Biao Yuan: Conceptualization, Methodology, Software, Modelling, Formal analysis, Visualization, Writing – original draft. He Wang: Conceptualization, Methodology, review & Editing. Yanjie Song: Conceptualization, Methodology, review & Editing. Ana Heitor:

Conceptualization, Methodology, review & Editing. Xiaohui Chen: Conceptualization, Methodology, review & Editing.

## Declaration of competing interest

The authors declare that he has no known competing financial interests or personal relationships that could have appeared to influence the work reported in this paper.

## Acknowledgements

The first author would like to acknowledge the financial support from the China Scholarship Council-University of Leeds Scholarships for his study at the University of Leeds. Acknowledgement to Horizon EU Marie Skłodowska-Curie Actions (MSCA, Grant Agreement ID: 101182689) through the GRID project for supporting this research.